\documentclass{article}

\usepackage[preprint]{neurips_2022}

\usepackage[utf8]{inputenc} 
\usepackage[T1]{fontenc}    
\usepackage{hyperref}       
\usepackage{url}            
\usepackage{booktabs}       
\usepackage{amsfonts}       
\usepackage{nicefrac}       
\usepackage{microtype}      
\usepackage{xcolor}         
\usepackage{graphicx}
\usepackage{url}
\usepackage{listings}
\usepackage{multirow}
\usepackage{subfigure}

\usepackage[utf8]{inputenc} 
\usepackage[T1]{fontenc}    
\usepackage{hyperref}       
\usepackage{url}            
\usepackage{booktabs}       
\usepackage{amsfonts}       
\usepackage{nicefrac}       
\usepackage{microtype}      
\usepackage{xcolor}         
\usepackage{epsfig}
\usepackage{graphicx}
\usepackage{subfigure}
\usepackage{booktabs}
\usepackage{caption}
\usepackage{array}
\usepackage{etoolbox}
\usepackage{amsmath} 
\usepackage{tabulary}
\usepackage{colortbl}
\usepackage{bbding}

\hypersetup{
    colorlinks=true,
    linkcolor=red,
    citecolor=cyan,
    filecolor=magenta,      
    urlcolor=cyan,
    }

\newtheorem{theorem}{Theorem}
\newtheorem{lemma}{Lemma}

\title{Divide to Adapt: Mitigating Confirmation Bias for Domain Adaptation of Black-Box Predictors}

\author{
  Jianfei Yang$^{1}$\thanks{Equal contribution. (yang0478@ntu.edu.sg, xiangyupeng@comp.nus.edu.sg)},\; Xiangyu Peng$^{2}$\footnotemark[1],\; Kai Wang$^{2,4}$,\; Zheng Zhu$^{3}$,\;\\ \textbf{Jiashi Feng}$^{4}$,\; \textbf{Lihua Xie}$^{1}$,\;  \textbf{Yang You}$^{2}$\thanks{Corresponding author (youy@comp.nus.edu.sg).}
   \\
  $^{1}$Nanyang Technological University, Singapore \quad
  $^{2}$National University of Singapore \\
  $^{3}$Tsinghua University \quad
  $^{4}$ByteDance\\
Code: \url{https://github.com/xyupeng/BETA}
}

\begin{document}

\maketitle

\begin{abstract}

Domain Adaptation of Black-box Predictors (DABP) aims to learn a model on an unlabeled target domain supervised by a black-box predictor trained on a source domain. It does not require access to both the source-domain data and the predictor parameters, thus addressing the data privacy and portability issues of standard domain adaptation. Existing DABP approaches mostly rely on model distillation from the black-box predictor, \emph{i.e.}, training the model with its noisy target-domain predictions, which however inevitably introduces the confirmation bias accumulated from the prediction noises. To mitigate such bias, we propose a new method, named BETA, to incorporate knowledge distillation and noisy label learning into one coherent framework. This is enabled by a new divide-to-adapt strategy. BETA divides the target domain into an easy-to-adapt subdomain with less noise and a hard-to-adapt subdomain. Then it deploys mutually-teaching twin networks to filter the predictor errors for each other and improve them progressively, from the easy to hard subdomains. As such, BETA effectively purifies the noisy labels and reduces error accumulation. We theoretically show that the target error of BETA is minimized by decreasing the noise ratio of the subdomains. Extensive experiments demonstrate BETA outperforms existing methods on all DABP benchmarks, and is even comparable with the standard domain adaptation methods that use the source-domain data.

\end{abstract}

\section{Introduction}
Unsupervised domain adaptation (UDA)~\cite{ganin15,long2016deep,saito2018maximum,zou2019consensus} aims to transfer knowledge from a labeled source domain to an unlabeled target domain and has wide applications~\cite{tzeng2015simultaneous,hoffman2018cycada,zou2021unsupervised,zhang2020collaborative}. 
However, UDA methods require to access the source-domain data, thus raising concerns about data privacy and portability issues. To sovle them, Domain Adaptation of Black-box Predictors (DABP)~\cite{liang2022dine} was introduced recently, which aims to learn a model with only the unlabeled target-domain data and a black-box predictor trained on the source domain, \textit{e.g.}, an API in the cloud, to avoid the privacy and safety issues from the leakage of data and model parameters.


A few efforts have been made to solve the DABP problem. One of them is to leverage knowledge distillation~\cite{hinton2015distilling} and train the target model to imitate predictions from the source predictor~\cite{liang2022dine}. Another one is to adopt learning with noisy labels (LNL) methods to select the clean samples from the noisy target-domain predictions for model training~\cite{zhang2021unsupervised}. Though inspiring, they have the following limitations. (i) Directly using the noisy pseudo labels for knowledge distillation inevitably leads to confirmation bias~\cite{tarvainen2017mean}, \textit{i.e.}, accumulated model prediction errors. (ii) The LNL-based methods only use a subset of the target dataset to train the model, which would limit the model's performance. (iii) Existing DABP methods lack theoretical justifications.

To address the aforementioned issues, this work proposes to mitigate the confirmation bias progressively while making full use of the target domain without forsaking any samples. This is achieved by a new divide-to-adapt strategy where the target domain is divided into an easy-to-adapt subdomain with less noise and a hard-to-adapt subdomain. The domain division is enabled by an observation: deep models tend to learn clean samples faster than noisy samples~\cite{arpit2017closer}. At the early stage of knowledge distillation in DABP, we find that there emerge two peaks in the loss distribution that can be fitted by a Gaussian Mixture Model (GMM)~\cite{arazo2019unsupervised} for domain division. Regarding the easy-to-adapt subdomain as a labeled set and the hard-to-adapt subdomain as an unlabeled set, we can leverage prevailing semi-supervised learning methods~\cite{berthelot2019mixmatch,sohn2020fixmatch} to solve the DABP problem. The new proposed strategy purifies the target domain to decrease the error accumulation while fully utilizing all the target dataset.

In this manner, this paper proposes \textbf{B}lack-Box Mod\textbf{E}l Adap\textbf{T}ation by Dom\textbf{A}in Division (\textbf{BETA}) that consists of two key modules to suppress the confirmation bias progressively. Specifically, as shown in Figure~\ref{fig:intuition}, the model is firstly warmed up by knowledge distillation from the source predictor. Then the domain division is performed based on a GMM that fits to the loss distribution. Leveraging the two subdomains, BETA adopts MixMatch~\cite{berthelot2019mixmatch} to fully utilize all target domain data for training. To further alleviate the confirmation bias, mutually-teaching twin networks are proposed in BETA where one network learns from the domain division of the other network, while the domain division is diverged by augmentation and Mixup techniques that enforce two networks to perform differently and filter the error for each other. From the theoretical analysis of BETA, the distribution discrepancy between two subdomains also matters and thus serves as an extra learning objective.

We make the following contributions. (i) We propose a novel BETA framework for the DABP problem that iteratively suppresses the error accumulation of model adaptation from the black-box source-domain predictor.
To the best of our knowledge, this is the first work that addresses the confirmation bias for DABP.
(ii) We theoretically show that the error of the target domain is bounded by the noise ratio of the hard-to-adapt subdomain, and empirically show that this error is minimized by BETA.
(iii) Extensive experiments demonstrate that our proposed BETA achieves state-of-the-art performance consistently on all benchmarks. Compared to the best baseline, our method leads to a significant improvement of \textbf{7.0}\% on the challenging benchmark, VisDA-17.
\section{Related Work}

\textbf{Unsupervised Domain Adaptation.}
Unsupervised domain adaptation aims to adapt a model from a labeled source domain to an unlabeled target domain. Early UDA methods rely on feature projection~\cite{pan2010domain,liang2018aggregating} and sample selection~\cite{huang2006correcting,sugiyama2007direct} for classic machine learning models. With the development of deep representation learning, deep domain adaptation methods yield surprising performances in challenging UDA scenarios. Inspired by two-sample test, discrepancy minimization of feature distributions~\cite{koniusz2017domain} is proposed to learn domain-invariant features~\cite{cui2020heuristic} based on statistic moment matching, such as Maximum Mean Discrepancy~\cite{TzengHZSD14,long2016deep} and covariance~\cite{sun2016deep}. Domain adversarial learning further employs a domain discriminator to achieve the same goal~\cite{ganin2016domain,long2017conditional,hoffman2018cycada,zou2019consensus,tzeng2017adversarial,yang2020mind,yang2021robust,xu2021partial} and achieves remarkable results. Other effective techniques for UDA include entropy minimization~\cite{grandvalet2005semi,cui2020towards,shu2018dirt}, contrastive learning~\cite{kang2019contrastive,huang2021model}, domain normalization~\cite{wang2019transferable,chang2019domain}, semantic alignment~\cite{liang2018aggregating,xie2018learning,deng2019cluster,yang2021advancing}, meta-learning~\cite{liu2020learning}, self-supervision~\cite{saito2020universal}, curriculum learning~\cite{zhang2017curriculum} and self-training~\cite{chen2020self,zou2018unsupervised,saito2017asymmetric}. Despite their effectiveness, they require the access to the source domain data and therefore invoke privacy and portability concerns.

\textbf{Unsupervised Model Adaptation and DABP.}
Without accessing the source domain, unsupervised model adaptation, \textit{i.e.}, source-free UDA, has attracted increasing attention since it loosens the assumption and benefits more practical scenarios \cite{guan2021domain}. Early research provides a theoretical analysis of hypothesis transfer learning~\cite{kuzborskij2013stability}, which motivates the existence of deep domain adaptation without source data~\cite{liang2020we,huang2021model,li2020model,dong2021confident}. Liang \textit{et al.} propose to train the feature extractor by self-supervised learning and mutual information maximization with the classifier frozen ~\cite{liang2020we}. Historical contrastive learning is developed to learn the target domain without forgetting the source hypothesis~\cite{huang2021model}.~\cite{li2020model} incorporates a GAN to generate images for adaptation. Nevertheless, these methods still require the model parameters. In this paper, we deal with a more challenging problem: \textit{only leveraging the labels from the model trained in the source domain for model adaptation}. Few works have been conducted in this field. \cite{zhang2021unsupervised} proposes a noisy label learning method by sample selection, while \cite{liang2022dine} using knowledge distillation with information maximization. Different from them, we propose to divide the target domain into two subdomains, which transforms black-box model adaptation into a partially-noisy semi-supervised learning task.

\textbf{Confirmation Bias in Semi-Supervised Learning.}
Confirmation bias refers to the noise accumulation when the model is trained using incorrect predictions for semi-supervised or unsupervised learning~\cite{tarvainen2017mean}. Such bias can cause the model to overfit the noisy feature space and then resist new changes~\cite{arazo2020pseudo}. In UDA, pseudo-labeling~\cite{saito2017asymmetric,gu2020spherical,morerio2020generative} and knowledge distillation~\cite{liang2020we,kundu2019adapt,zhou2020domain} are effective techniques but can be degraded due to confirmation bias. Especially for the transfer task with a distant domain, the pseudo labels for the target domain are very noisy and deteriorate the subsequent epochs of training. To alleviate the confirmation bias, co-training~\cite{qiao2018deep,li2019dividemix} employs two or more networks simultaneously to obtain complementary supervision from each other. The Mixup enforces the model to favor simple linear behavior in-between training examples by interpolation consistency~\cite{zhang2018mixup}, and data-augmented unlabeled examples~\cite{cubuk2019autoaugment} further diverge the pairs of Mixup to further eliminate the confirmation bias~\cite{berthelot2019mixmatch}. Our paper proposes mutually-teaching twin networks with subdomain augmentation to suppress the confirmation bias. To the best of our knowledge, this is the first work that formulates and addresses the confirmation bias for DABP.

\section{Methodology}
The idea of our proposed BETA is to mitigate the confirmation bias for DABP by dividing the target domain into two subdomains with different adaptation difficulties. As shown in Figure~\ref{fig:intuition}, BETA relies on two designs to suppress error accumulation, including a domain division module that purifies the target domain into a cleaner subdomain and transfers DABP to a semi-supervised learning task, and a two-networks mechanism (\textit{i.e.}, mutually-teaching twin networks) that further diminishes the self-training errors by information exchange. We firstly introduce the problem formulation and the key modules, and then make the algorithmic instantiation with more details.

\subsection{Problem Formulation}
For domain adaptation of black-box predictors, the model has access to a black-box predictor $h_s$ trained by a source domain $\{(\mathbf{x}^s_i, y^s_i)\}_{i=1}^{N_s}$ with $N_s$ labeled samples where $\mathbf{x}^s_i\in\mathcal{X}_s,y^s_i\in\mathcal{Y}_s$, and an unlabeled target domain $\{\mathbf{x}^t_i\}_{i=1}^{N_t}$ with $N_t$ unlabeled samples where $\mathbf{x}^t_i\in\mathcal{X}_t$. Assume that the label spaces are the same across two domains, \textit{i.e.}, $\mathcal{Y}_s=\mathcal{Y}_t$, while the inputs data have different distributions, i.e. $\mathbf{x}^s_i \sim \mathcal{D}_S$ and $\mathbf{x}^t_i \sim \mathcal{D}_T$. In other words, there exist a \textit{domain shift}~\cite{ben2007analysis} between $\mathcal{D}_S$ and $\mathcal{D}_T$. The objective is to learn a mapping model $\mathcal{X}_t\rightarrow \mathcal{Y}_t$. Different from standard UDA~\cite{pan2010survey,TzengHZSD14,long2016deep}, DABP prohibits the model from accessing the source-domain data $\mathcal{X}_s,\mathcal{Y}_s$, and the parameters of the source model $h_s$. Only a black-box predictor trained on the source domain, \textit{i.e.}, an API, is available. Confront these constraints, we can only resort to the hard predictions of the target domain from the source predictor, \emph{i.e.}, $\tilde{\mathcal{Y}}_t=h_s(\mathcal{X}_t)$, in the DABP setting. 

\subsection{Domain Division}
\begin{figure*}[t]
	\centering
	\includegraphics[width=0.95\textwidth]{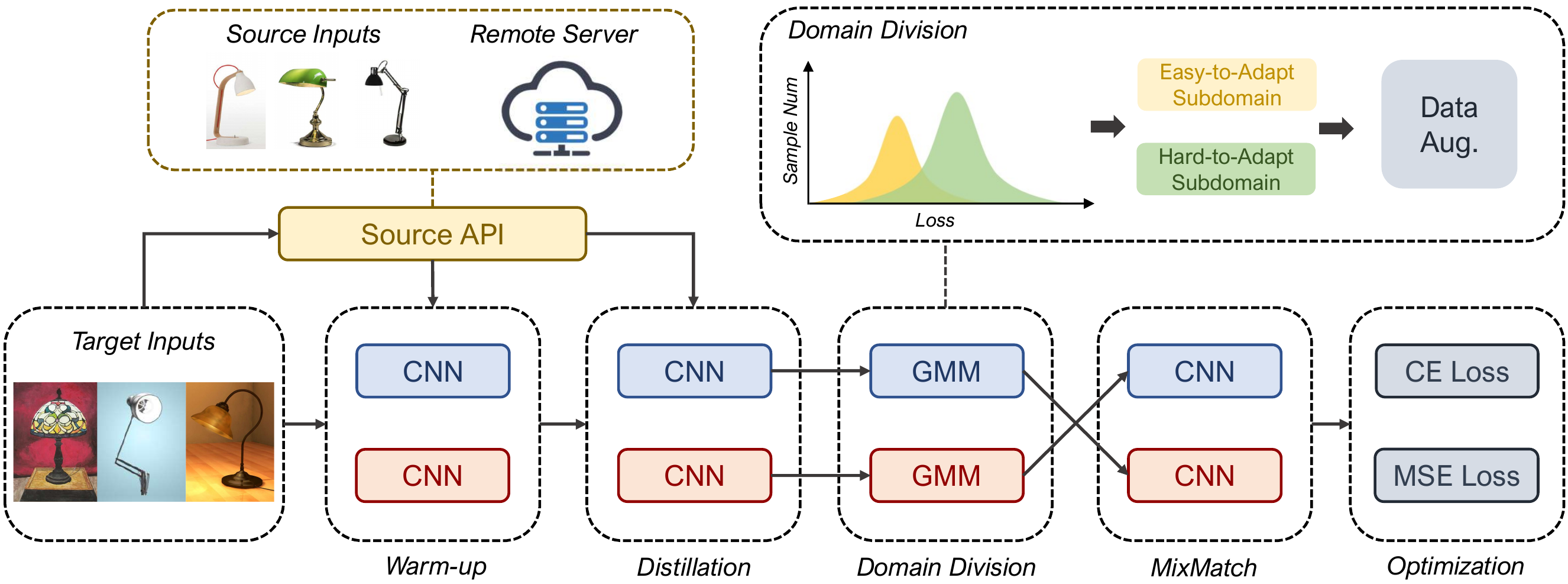}
	\caption{BETA has mutually-teaching twin networks initialized by the predictions from the source API. Then the target domain is divided into a clean subdomain and a noisy one. The two networks utilize each other's domain division for semi-supervised learning (\textit{e.g.}, MixMatch).}
	\label{fig:intuition}
	\vspace{-3mm}
\end{figure*}

Different from the strategy of directly utilizing $\mathcal{X}_t,\tilde{\mathcal{Y}}_t$ for knowledge distillation~\cite{liang2022dine}, we propose to divide the target domain $\mathcal{X}_t$ into an \textit{easy-to-adapt} subdomain $\mathcal{X}_t^e\sim \mathcal{D}_e$ and a \textit{hard-to-adapt} subdomain $\mathcal{X}_t^h\sim \mathcal{D}_h$, with $\mathcal{X}_t=\mathcal{X}_e \cup \mathcal{X}_h$. Previous studies show that deep models are prone to fitting clean examples faster than noisy examples~\cite{arpit2017closer,li2019dividemix,wang2022reliable}, which inspires us to divide the domain by loss values at the early stage of training. Clean examples are usually highly-confident samples assigned to $\mathcal{X}_e$ while noisy examples are uncertain examples assigned to $\mathcal{X}_h$. Based on this observation, we first warm up the network, \textit{e.g.}, a CNN, for several epochs, and then obtain the loss distribution by calculating the per-sample cross-entropy loss for a $K$-way classification problem as
\begin{equation}
    \mathcal{L}_{ce}(\mathbf{x}_i^t)=-\sum_{k=1}^K \tilde{y}_i^k\log (h_t^k(\mathbf{x}_i^t)),
\end{equation}
where $h_t^k$ is the softmax probability for class $k$ from the target model. As shown in Figure~\ref{fig:intuition}, the loss distribution appears to be bimodal and two peaks indicate the clean and noisy clusters, which can be fitted by a GMM~\cite{li2019dividemix}. In noisy label learning, the clean and noisy subset division is achieved by a Beta Mixture Model (BMM)~\cite{arazo2019unsupervised}. However, in DABP, the noisy pseudo labels obtained by $h_s$ are dominated by asymmetric noise~\cite{kim2021fine}, \textit{i.e.}, the noisy samples that do not follow a uniform distribution. In this case, the BMM leads to undesirable flat distributions and cannot work effectively for our task~\cite{song2022learning}. Asymmetric noise in $\tilde{\mathcal{Y}}_t$ also causes the model to perform confidently and generate near-zero losses, which hinders the domain division of GMM. To better fit the losses of the target domain with asymmetric noise, the negative entropy is used as a regularizer in the warm-up phase, defined as:
\begin{equation}
    \mathcal{L}_{ne}=\sum_{k=1}^K h_t^k(\mathbf{x}_i^t)\log (h_t^k(\mathbf{x}_i^t)).
\end{equation}

After fitting the loss distribution to a two-component GMM via the Expectation-Maximization algorithm, the clean probability $\varrho_i^c$ is equivalent to the posterior probability $p(c|l_i(\mathbf{x}_i^t))$ where $c$ is the Gaussian component with smaller loss. Then the clean and noisy subdomains are divided by setting a threshold $\tau$ based on the clean probabilities:
\begin{align}
    & \mathcal{X}_e=\{ (\mathbf{x}_i,\tilde{y}_i)|(\mathbf{x}_i,\tilde{y}_i)\in (\mathcal{X}_t,\tilde{\mathcal{Y}}_t), \varrho_i^c\ge \tau \}, \\
    & \mathcal{X}_h=\{(\mathbf{x}_i,\tilde{p}_i)|\mathbf{x}_i\in \mathcal{X}_t, \varrho_i^c< \tau \}, 
\end{align}
where $\tilde{p}_i=h_s(\mathbf{x}_i^t)$ is the softmax probabilities. Intuitively, the clean subdomain consists of \textit{easy-to-adapt} samples, while the noisy subdomain consists of \textit{hard-to-adapt} samples. The semi-supervised learning methods~\cite{berthelot2019mixmatch} can be directly applied with $\mathcal{X}_e$ used as the labeled set and $\mathcal{X}_h$ as the unlabeled set. Compared to sample selection~\cite{zhang2021unsupervised} and single distillation~\cite{liang2022dine}, domain division enables the utilization of all accessible samples by semi-supervised learning and dilutes the risk of the confirmation bias by leveraging cleaner signals of supervision for model adaptation.

\subsection{Mutually-teaching Twin Networks with Subdomain Augmentation}
The easy-to-adapt subdomain is purified by domain division but still has inevitable wrong labels. Overfitting to these wrong labels enforces the model to generate fallacious low losses for domain division, and hence accumulates the wrong predictions iteratively, which is the confirmation bias. Apart from domain division, we propose Mutually-teaching Twin Networks (MTN) to further mitigate such bias, inspired by the two-networks design in~\cite{qiao2018deep,li2019dividemix} where the confirmation bias of self-training can be diminished by training two networks to decontaminate the noise for each other. Specifically, we employ two identical networks initialized independently, where one network performs semi-supervised learning according to the domain division and pseudo labels of the other network. In this fashion, two networks are trained mutually and receive extra supervision to filter the error.

In BETA, we further revamp this design by subdomain augmentation to increase the divergence of domain division, enabling two networks to obtain sufficiently different supervisions from each other.
Suppose that two networks $h_t^{\theta_1},h_t^{\theta_2}$ where $\theta_1,\theta_2$ are parameters generate two sets of domain division $\{\mathcal{X}_e^1,\mathcal{X}_h^1\}$ and $\{\mathcal{X}_e^2,\mathcal{X}_h^2\}$, respectively. We take $h_t^{\theta_2}$ and $\{\mathcal{X}_e^1,\mathcal{X}_h^1\}$ for example. Two augmentation strategies are tailored: the weak augmentation (\textit{e.g.}, random cropping and flipping), and the strong augmentation (\textit{i.e.}, RandAugment~\cite{cubuk2020randaugment} and AutoAugment~\cite{cubuk2019autoaugment}. The samples from the easy-to-adapt subdomain are mostly correct, so we augment them using two strategies and obtain their soft pseudo labels by the convex combination of averaging all augmentations and the pseudo label according to the clean probability $\varrho_i^c$. Whereas, the hard-to-adapt subdomain is noisy, so we only apply the weak augmentation to update their pseudo labels but use strong augmentations in the subsequent learning phase. Furthermore, we employ the co-guessing strategy~\cite{li2019dividemix} to refine the pseudo labels for $\mathcal{X}_h$. The refined subdomains are derived as:
\begin{align}
    & \widehat{\mathcal{X}}_e^1= \big \{ (\mathbf{x}_i,\tilde{y}'_i)|
    \tilde{y}'_i=\varrho_i^c \tilde{y}_i+(1-\varrho_i^c){\scriptsize \frac{1}{M}\sum_{m=1}^M}[ h_t^{\theta_2}(A^m_{w/s}(\mathbf{x}_i))],
    (\mathbf{x}_i,\tilde{y}_i)\in \mathcal{X}_e^1 \big \}, \\
    & \widehat{\mathcal{X}}_h^1= \big \{(\mathbf{x}_i,\tilde{p}'_i)| 
    \tilde{p}'_i=\frac{1}{2M}\sum_{m=1}^{2M}[h_t^{\theta_1}(A^m(\mathbf{x}_i))+h_t^{\theta_2}(A^m_w(\mathbf{x}_i))],
    (\mathbf{x}_i,\tilde{p}_i)\in \mathcal{X}_h^1
    \big \}, 
\end{align}
where $A^m_{w/s}(\cdot)$ denotes the $m$-th weak and strong augmentation function, and $M$ denotes the total number of augmentation views.

\subsection{Algorithmic Instantiation}
After the warm-up, domain division, and subdomain augmentation, we detail the algorithmic choices of other modules and the learning objectives of our framework.

\textbf{Hard knowledge distillation.} For each epoch, we first perform knowledge distillation from the predictions of the source model $h_s$ by the relative entropy, \textit{i.e.}, the Kullback–Leibler divergence
\begin{equation}
    \mathcal{L}_{kd}=\mathbb{E}_{\mathbf{x}_t\in\mathcal{X}_t}D(\tilde{y}_t||h_t(\mathbf{x}_t)),
\end{equation}
where $D_{KL}(\cdot||\cdot)$ denotes the KL-divergence, and the pseudo label $\tilde{y}_t$ is obtained by the EMA prediction of $h_s(\mathbf{x}_t)$. Different from DINE~\cite{liang2022dine} that uses source model probabilities, we only leverage the hard pseudo labels that are more ubiquitous for API services.

\textbf{Mutual information maximization.} To circumvent the model to show partiality for categories with more samples during knowledge distillation, we maximize the mutual information as
\begin{equation}
    \mathcal{L}_{mi}=H(\mathbb{E}_{x\in\mathcal{X}_t} h_t(x))-\mathbb{E}_{x\in\mathcal{X}_t} H(h_t(x)),
\end{equation}
where $H(\cdot)$ denotes the information entropy. This loss works jointly with $\mathcal{L}_{kd}$. Besides, after the semi-supervised learning, we use this loss to fine-tune the model to enforce the model to comply with the cluster assumption~\cite{shu2018dirt,liang2022dine,grandvalet2005semi}.

\textbf{Domain division enabled semi-supervised learning.} We choose MixMatch~\cite{berthelot2019mixmatch} as the semi-supervised learning method since it includes a mix-up procedure~\cite{zhang2018mixup} that can further diverge the two networks while refraining from overfitting. The mixed sets $\Ddot{\mathcal{X}}_e,\Ddot{\mathcal{X}}_h$ are obtained by $\Ddot{\mathcal{X}}_e=\text{Mixup}(\widehat{\mathcal{X}}_e,\widehat{\mathcal{X}}_e\cup\widehat{\mathcal{X}}_h)$ and $\Ddot{\mathcal{X}}_h=\text{Mixup}(\widehat{\mathcal{X}}_h,\widehat{\mathcal{X}}_e\cup\widehat{\mathcal{X}}_h)$. Then the loss functions is written as
\begin{equation}
    \mathcal{L}_{dd}=\mathcal{L}_{ce}(\Ddot{\mathcal{X}}_e)+\mathcal{L}_{mse}(\Ddot{\mathcal{X}}_h)+\mathcal{L}_{reg},
\end{equation}
where $\mathcal{L}_{ce}$ denotes the cross-entropy loss, $\mathcal{L}_{mse}$ denotes the mean squared error, and the regularizer $\mathcal{L}_{reg}$ uses a uniform distribution $\pi_k$ to eliminate the effect of class imbalance, written as
\begin{equation}
    \mathcal{L}_{reg}=\sum_k\pi_k\log \bigg(\pi_k\bigg/\frac{1}{|\Ddot{\mathcal{X}}_e|+|\Ddot{\mathcal{X}}_h|}\sum_{\mathbf{x}\in \Ddot{\mathcal{X}}_e+\Ddot{\mathcal{X}}_h}h_t(\mathbf{x})\bigg).
\end{equation}

\textbf{Subdomain alignment.} We assume that there exists a distribution discrepancy between the easy-to-adapt and the hard-to-adapt subdomains, which leading to the performance gap between them. Regarding this gap, we add an adversarial regularizer by introducing a domain discriminator $\Omega(\cdot)$:
\begin{equation}
    \mathcal{L}_{adv}=\mathbb{E}_{\mathbf{x}\in \Ddot{\mathcal{X}}_e}\log \big(\Omega(h_t(x))\big)+
    \mathbb{E}_{\mathbf{x}\in \Ddot{\mathcal{X}}_h}\log\big(1-\Omega(h_t(x))\big).
\end{equation}

\textbf{Overall objectives.}
Summarizing all the losses, the overall objectives are formulated as
\begin{equation}
    \mathcal{L}=(\underbrace{\mathcal{L}_{kd}-\mathcal{L}_{im}}_{step\, 1})+(\underbrace{\mathcal{L}_{dd}-\gamma \mathcal{L}_{adv}}_{step\,2}),
\end{equation}
where $\gamma$ is a hyper-parameter that is empirically set to 0.1. In step 1, we perform distillation for two networks independently to form tight clusters by maximizing mutual information, while in step 2, the proposed BETA revises their predictions by mitigating the confirmation bias in a synergistic manner. The domain division is performed between two steps.

\subsection{Theoretical Justifications}\label{sec:theory}
Existing theories on UDA error bound~\cite{ben2007analysis,ben2010theory,ganin2016domain,long2017conditional} are based on the source-domain data, so they are not applicable to DABP models~\cite{liang2022dine,zhang2021unsupervised}, which hinders  understanding of these models. To better explain why BETA contributes to DABP, we derive an error bound based on the existing UDA theories~\cite{ben2007analysis,ben2010theory}. Let $h$ denote a hypothesis, $y_e,y_h$ and $\hat{y}_e,\hat{y}_h$ denote the ground truth labels and the pseudo labels of $\mathcal{X}_e, \mathcal{X}_h$, respectively. As BETA is trained on a mixture of the two subdomains with pseudo labels, the error of BETA can be formulated as a convex combination of the errors of the easy-to-adapt subdomain and the hard-to-adapt subdomain:
\begin{equation}
    \epsilon_\alpha(h) = \alpha \epsilon_e(h,\hat{y}_e)+(1-\alpha)\epsilon_h(h,\hat{y}_h),
\end{equation}
where $\alpha$ is the trade-off hyper-parameter, and $\epsilon_e(h,\hat{y}_e),\epsilon_h(h,\hat{y}_h)$ represents the expected error of the two subdomains. We derive an upper bound of how the error $\epsilon_\alpha(h)$ is close to an oracle error of the target domain $\epsilon_t(h,y_t)$ where $y_t$ is the ground truth labels of the target domain.

\begin{theorem}\label{theorem}
Let $h$ be a hypothesis in class $\mathcal{H}$. Then
\begin{equation}
|\epsilon_\alpha(h)-\epsilon_t(h,y_t)|\leq \alpha(d_{\mathcal{H}\triangle\mathcal{H}}(\mathcal{D}_e,\mathcal{D}_h)+\lambda+\hat{\lambda})+\rho_h,
\end{equation}
where the ideal risk is the combined error of the ideal joint hypothesis $\lambda=\epsilon_e(h^*)+\epsilon_h(h^*)$, the distribution discrepancy $d_\mathcal{\mathcal{H}\triangle\mathcal{H}}(\mathcal{D}_e,\mathcal{D}_h)=2\sup_{h,h' \in \mathcal{H}} |\mathbb{E}_{x\sim \mathcal{D}_c}[h(x)\ne h'(x)] - \mathbb{E}_{x\sim \mathcal{D}_n}[h(x)\ne h'(x)]|$, and $\rho_h$ denote the pseudo label rate of $\hat{y}_h$. The ideal joint hypothesis is given by $h^*=\arg \min_{h\in\mathcal{H}}(\epsilon_e(h)+\epsilon_h(h))$, deriving the ideal risk $\lambda=\epsilon_e(h^*)+\epsilon_h(h^*)$ and the pseudo risk $\hat{\lambda}=\epsilon_e(h^*,\hat{y}_e)+\epsilon_h(h^*,\hat{y}_h)$.
\end{theorem}

In the above theorem, the error is bounded by the distribution discrepancy between two subdomains, the noise ratio of $\mathcal{X}_h$, and the risks. The ideal risk $\lambda$ is neglectly small~\cite{ganin2016domain}, and the pseudo risk $\hat{\lambda}$ is bounded by $\rho_h$ as shown in the appendix. Hence, the subdomain discrepancy and $\rho_h$ dominate the error bound. Empirical results show that $d_\mathcal{\mathcal{H}\triangle\mathcal{H}}(\mathcal{D}_e,\mathcal{D}_h)$ is usually small for the two subdomains, and $\rho_h$ keeps dropping during training as shown in Figure~\ref{fig:confirmation-bias}, which tightens the upper bound consequently. The proof with detailed analytics is in the appendix.

\section{Experiments}
\subsection{Setup}\label{sec:exp-setting}
\textbf{Datasets.} \textbf{Office-31}~\cite{saenko2010adapting} is the most common benchmark for UDA, which consists of three domains (\underline{\textbf{A}}mazon, \underline{\textbf{W}}ebcam, \underline{\textbf{D}}SLR) in 31 categories.
\textbf{Office-Home}~\cite{venkateswara2017deep} consists of four domains (\underline{\textbf{Ar}}t, \underline{\textbf{Cl}}ipart, \underline{\textbf{Pr}}oduct, \underline{\textbf{Re}}al World) in 65 categories, and the distant domain shifts render it more challenging.
\textbf{VisDA-17}~\cite{peng2017visda} is a large-scale benchmark for synthetic-to-real object recognition, with a source domain with 152k synthetic images and a target domain with 55k real images from Microsoft COCO.

\textbf{Implementation details.} We implement our method via PyTorch~\cite{paszke2019pytorch}, and report the average accuracies among three runs. To show the capacity of handling the confirmation bias, we further report the average accuracies across hard tasks whose source-only accuracies are below 65\% (\textbf{H. Avg.}). We employ ResNet-50 for Office-31 and Office-Home, and ResNet-101 for VisDA-17 as the backbones~\cite{he2016deep}, and add a new MLP-based classifier, which is commonly used in existing UDA works~\cite{long2016deep,chen2019transferability,liang2022dine,saito2018maximum}. The domain discriminator consists of fully-connected layers (2048-256-2) that perform a binary subdomain classification~\cite{long2017conditional}. The ImageNet pre-trained model is utilized as initialization. The model is optimized by mini-batch SGD with the learning rate of 1e-3 for CNN layers and 1e-2 for the MLP classifier. Following DINE~\cite{liang2022dine}, we use the suggested training strategies including the momentum (0.9), batch size (64), and weight decay (1e-3). The number of epochs for warm-up is empirically set to 3, and the training epoch is 50 except 10 for VisDA-17. The hyper-parameters of MixMatch are kept as same as the original paper~\cite{berthelot2019mixmatch}, attached in the appendix. As two networks of MTN perform similarly, we report the accuracy of the first network.

\textbf{Baselines.} For a fair comparison, we follow the protocol and training strategy for the source domain in DINE~\cite{liang2022dine}, and compare our BETA with state-of-the-art DABP methods. Specifically, \textbf{LNL-KL}~\cite{zhang2021unsupervised} and \textbf{LNL-OT}~\cite{asano2019self} are noisy label learning methods with KL divergence and optimal transport, respectively. \textbf{HD-SHOT} and \textbf{SD-SHOT} obtain the model using pseudo labels and apply SHOT~\cite{liang2020we} by self-training and the weighted cross-entropy loss, respectively. We also show state-of-the-art standard UDA methods for comparison, including CDAN~\cite{long2017conditional}, MDD~\cite{zhang2019bridging}, BSP~\cite{chen2019transferability}, CKB~\cite{luo2021conditional}, CST~\cite{liu2021cycle}, SAFN~\cite{xu2019larger}, DTA~\cite{lee2019drop}, ATDOC~\cite{liang2021domain}, MCC~\cite{jin2020minimum}, BA$^3$US~\cite{liang2020balanced} and JUMBOT~\cite{fatras2021unbalanced}.


\begin{table}[tp]
    \centering
    \footnotesize
    \caption{Accuracies (\%) on Office-31 for black-box model adaptation. {\colorbox[HTML]{DAE8FC}{H. Avg.}} denotes the average accuracy of the hard tasks whose source-only accuracies are below 65\%.}
    \label{tab:office-31}
    \scalebox{0.8}{
        \begin{tabular}{
        l|c|c|cccccc
        >{\columncolor[HTML]{FFF2CC}}c 
        >{\columncolor[HTML]{DAE8FC}}c }
        \toprule
        Method   & Publication &   DABP    & A→D           & A→W           & \cellcolor[HTML]{DAE8FC}D→A & D→W           & \cellcolor[HTML]{DAE8FC}W→A & W→D         & Avg.          & H. Avg.     \\ \midrule
        ResNet-50 \cite{he2016deep} & -  &  -   & 79.9          & 76.6          & 56.4                        & 92.8          & 60.9                        & 98.5        & 77.5          & 58.7          \\
        LNL-OT \cite{asano2019self} & ICLR-19  &  \Checkmark    & 88.8          & 85.5          & 64.6                        & 95.1          & 66.7                        & 98.7        & 83.2          & 65.7          \\
        LNL-KL \cite{zhang2021unsupervised} & BMVC-21 &  \Checkmark     & 89.4          & 86.8          & 65.1                        & 94.8          & 67.1                        & 98.7        & 83.6          & 66.1          \\
        HD-SHOT \cite{liang2021source} & TPAMI-21 &  \Checkmark    & 86.5          & 83.1          & 66.1                        & 95.1          & 68.9                        & 98.1        & 83.0            & 67.5          \\
        SD-SHOT \cite{liang2021source} & TPAMI-21 &  \Checkmark    & 89.2          & 83.7          & 67.9                        & 95.3          & 71.1                        & 97.1        & 84.1          & 69.5          \\
        DINE \cite{liang2022dine} & CVPR-22  &  \Checkmark      & 91.6          & 86.8          & 72.2                        & \textbf{96.2}          & 73.3                        & 98.6        & 86.4          & 72.8          \\ \midrule
        \textbf{BETA} (Ours) & - &  \Checkmark & \textbf{93.6} & \textbf{88.3} & \textbf{76.1}               & 95.5 & \textbf{76.5}               & \textbf{99.0} & \textbf{88.2} & \textbf{76.3}\\ \midrule
        BSP+DANN \cite{chen2019transferability} & ICML-19 &  \XSolidBrush & 93.0 & 93.3 & 73.6                        & 98.2 & 72.6                        & 100.0 & 88.5 & 73.1    \\
        MDD \cite{zhang2019bridging} & ICML-19  &  \XSolidBrush                          & 93.5 & 94.5 & 74.6                        & 98.4 & 72.2                        & 100.0 & 88.9 & 73.4    \\
        ATDOC \cite{liang2021domain} & CVPR-21  &  \XSolidBrush                        & 94.4 & 94.3 & 75.6                        & 98.9 & 75.2                        & 99.6  & 89.7 & 75.4   \\
        \bottomrule
        \end{tabular}
    }
    \vspace{-3mm}
\end{table}

\begin{table}[t]
    \centering
    \footnotesize
    \caption{Accuracies (\%) on Office-Home for black-box model adaptation. (`:' denotes `transfer to')}\label{tab:office-home}
    \scalebox{0.69}{
    \begin{tabular}{l|c|cccccccccccccc}
    \toprule
    Method  & DABP  & \cellcolor[HTML]{DAE8FC}Ar:Cl & Ar:Pr         & Ar:Re         & \cellcolor[HTML]{DAE8FC}Cl:Ar & \cellcolor[HTML]{DAE8FC}Cl:Pr & Cl:Re         & \cellcolor[HTML]{DAE8FC}Pr:Ar & \cellcolor[HTML]{DAE8FC}Pr:Cl & Pr:Re         & Re:Ar         & \cellcolor[HTML]{DAE8FC}Re:Cl & Re:Pr         & \cellcolor[HTML]{FFF2CC}Avg.          & \cellcolor[HTML]{DAE8FC}H. Avg.     \\ \midrule
    \rowcolor[HTML]{FFFFFF} 
    ResNet-50 \cite{he2016deep} & - & 44.1                          & 66.9          & 74.2          & 54.5                          & 63.3                          & 66.1          & 52.8                          & 41.2                          & 73.2          & 66.1          & 46.7                          & 77.5          & \cellcolor[HTML]{FFF2CC}60.6          & \cellcolor[HTML]{DAE8FC}50.4          \\
    \rowcolor[HTML]{FFFFFF} 
    LNL-OT \cite{asano2019self} &  \Checkmark  & 49.1                          & 71.7          & 77.3          & 60.2                          & 68.7                          & 73.1          & 57.0                          & 46.5                          & 76.8          & 67.1          & 52.3                          & 79.5          & \cellcolor[HTML]{FFF2CC}64.9          & \cellcolor[HTML]{DAE8FC}55.6          \\
    \rowcolor[HTML]{FFFFFF} 
    LNL-KL \cite{zhang2021unsupervised} &  \Checkmark  & 49.0                          & 71.5          & 77.1          & 59.0                          & 68.7                          & 72.9          & 56.4                          & 46.9                          & 76.6          & 66.2          & 52.3                          & 79.1          & \cellcolor[HTML]{FFF2CC}64.6          & \cellcolor[HTML]{DAE8FC}55.4          \\
    \rowcolor[HTML]{FFFFFF} 
    HD-SHOT \cite{liang2021source} &  \Checkmark & 48.6                          & 72.8          & 77.0          & 60.7                          & 70.0                          & 73.2          & 56.6                          & 47.0                          & 76.7          & 67.5          & 52.6                          & 80.2          & \cellcolor[HTML]{FFF2CC}65.3          & \cellcolor[HTML]{DAE8FC}55.9          \\
    \rowcolor[HTML]{FFFFFF} 
    SD-SHOT \cite{liang2021source} &  \Checkmark & 50.1                          & 75.0          & 78.8          & 63.2                          & 72.9                          & 76.4          & 60.0                          & 48.0                          & 79.4          & 69.2          & 54.2                          & 81.6          & \cellcolor[HTML]{FFF2CC}67.4          & \cellcolor[HTML]{DAE8FC}58.1          \\
    \rowcolor[HTML]{FFFFFF} 
    \rowcolor[HTML]{FFFFFF} 
    DINE  \cite{liang2022dine} &  \Checkmark   & 52.2                          & 78.4          & 81.3          & 65.3                          & 76.6                          & 78.7          & 62.7                          & 49.6                          & 82.2          & 69.8          & 55.8                          & 84.2          & \cellcolor[HTML]{FFF2CC}69.7          & \cellcolor[HTML]{DAE8FC}60.4          \\ \midrule
    \rowcolor[HTML]{FFFFFF} 
    \textbf{BETA} (Ours) &  \Checkmark     & \textbf{57.2}                 & \textbf{78.5} & \textbf{82.1} & \textbf{68.0}                 & \textbf{78.6}                 & \textbf{79.7} & \textbf{67.5}                 & \textbf{56.0}                 & \textbf{83.0} & \textbf{71.9} & \textbf{58.9}                 & \textbf{84.2} & \cellcolor[HTML]{FFF2CC}\textbf{72.1} & \cellcolor[HTML]{DAE8FC}\textbf{64.4} \\ \midrule
    \rowcolor[HTML]{FFFFFF} 
    CDAN+E \cite{long2017conditional} &  \XSolidBrush  & 50.7                          & 70.6          & 76.0          & 57.6                          & 70.0                          & 70.0          & 57.4                          & 50.9                          & 77.3          & 70.9          & 56.7                          & 81.6          & \cellcolor[HTML]{FFF2CC}65.8          & \cellcolor[HTML]{DAE8FC}57.2          \\
    \rowcolor[HTML]{FFFFFF} 
    BSP+CDAN \cite{chen2019transferability} &  \XSolidBrush & 52.0                          & 68.6          & 76.1          & 58.0                          & 70.3                          & 70.2          & 58.6                          & 50.2                          & 77.6          & 72.2          & 59.3                          & 81.9          & \cellcolor[HTML]{FFF2CC}66.3          & \cellcolor[HTML]{DAE8FC}58.1          \\
    \rowcolor[HTML]{FFFFFF} 
    MDD \cite{zhang2019bridging}  &  \XSolidBrush   & 54.9                          & 73.7          & 77.8          & 60.0                          & 71.4                          & 71.8          & 61.2                          & 53.6                          & 78.1          & 72.5          & 60.2                          & 82.3          & \cellcolor[HTML]{FFF2CC}68.1          & \cellcolor[HTML]{DAE8FC}60.2          \\
    \rowcolor[HTML]{FFFFFF} 
    CKB \cite{luo2021conditional} &  \XSolidBrush    & 54.7                          & 74.4          & 77.1          & 63.7                          & 72.2                          & 71.8          & 64.1                          & 51.7                          & 78.4          & 73.1          & 58.0                          & 82.4          & \cellcolor[HTML]{FFF2CC}68.5          & \cellcolor[HTML]{DAE8FC}60.7          \\
    \rowcolor[HTML]{FFFFFF} 
    CST \cite{liu2021cycle} &  \XSolidBrush    & 59.0                          & 79.6          & 83.4          & 68.4                          & 77.1                          & 76.7          & 68.9                          & 56.4                          & 83.0          & 75.3          & 62.2                          & 85.1          & \cellcolor[HTML]{FFF2CC}73.0          & \cellcolor[HTML]{DAE8FC}65.3 \\ \bottomrule      
    \end{tabular}
    }
    \vspace{-3mm}
\end{table}
\begin{table}[t]
\centering
    \caption{Accuracies (\%) on VisDA-17 for black-box model adaptation.}\label{tab:visda-17}
    \scalebox{0.65}{
\begin{tabular}{
l|c|cccccccccccc
>{\columncolor[HTML]{FFF2CC}}c 
>{\columncolor[HTML]{DAE8FC}}c }
\toprule
Method   & DABP   & \cellcolor[HTML]{DAE8FC}plane & \cellcolor[HTML]{DAE8FC}bcycl & \cellcolor[HTML]{DAE8FC}bus & car           & horse         & \cellcolor[HTML]{DAE8FC}knife & mcycle        & \cellcolor[HTML]{DAE8FC}person & \cellcolor[HTML]{DAE8FC}plant         & \cellcolor[HTML]{DAE8FC}sktbrd & train         & \cellcolor[HTML]{DAE8FC}truck & Per-class     & H. Avg.       \\ \midrule
ResNet-101 \cite{he2016deep} & - & 64.3                          & 24.6                          & 47.9                        & 75.3          & 69.6          & 8.5                           & 79.0          & 31.6                           & 64.4          & 31.0                           & 81.4          & 9.2                           & 48.9          & 35.2          \\
LNL-OT  \cite{asano2019self} &  \Checkmark    & 82.6                          & 84.1                          & 76.2                        & 44.8          & 90.8          & 39.1                          & 76.7          & 72.0                           & 82.6          & 81.2                           & 82.7          & 50.6                          & 72.0          & 71.1          \\
LNL-KL \cite{zhang2021unsupervised} &  \Checkmark     & 82.7                          & 83.4                          & 76.7                        & 44.9          & 90.9          & 38.5                          & 78.4          & 71.6                           & 82.4          & 80.3                           & 82.9          & 50.4                          & 71.9          & 70.8          \\
HD-SHOT \cite{liang2021source}  &  \Checkmark   & 75.8                          & 85.8                          & 78.0                        & 43.1          & 92.0          & 41.0                          & 79.9          & 78.1                           & 84.2          & 86.4                           & 81.0          & \textbf{65.5}                          & 74.2          & 74.4          \\
SD-SHOT  \cite{liang2021source} &  \Checkmark   & 79.1                          & 85.8                          & 77.2                        & 43.4          & 91.6          & 41.0                          & 80.0          & 78.3                           & 84.7          & 86.8                           & 81.1          & 65.1                          & 74.5          & 74.8          \\
DINE  \cite{liang2022dine} &  \Checkmark      & 81.4                          & 86.7                          & 77.9                        & 55.1          & 92.2          & 34.6                          & 80.8          & 79.9                           & 87.3          & 87.9                           & 84.3          & 58.7                          & 75.6          & 74.3          \\ \midrule
\textbf{BETA} (Ours) &  \Checkmark        & \textbf{96.2}                 & \textbf{83.9}                 & \textbf{82.3}               & \textbf{71.0} & \textbf{95.3} & \textbf{73.1}                 & \textbf{88.4} & \textbf{80.6}                  & \textbf{95.5} & \textbf{90.9}                  & \textbf{88.3} & 45.1                 & \textbf{82.6} & \textbf{81.0} \\ \midrule
SAFN  \cite{xu2019larger}  &  \XSolidBrush      & 93.6                          & 61.3                          & 84.1                        & 70.6          & 94.1          & 79.0                          & 91.8          & 79.6                           & 89.9          & 55.6                           & 89.0          & 24.4                          & 76.1          & 70.9          \\
CDAN+E  \cite{long2017conditional} &  \XSolidBrush     & 94.3                          & 60.8                          & 79.9                        & 72.7          & 89.5          & 86.8                          & 92.4          & 81.4                           & 88.9          & 72.9                           & 87.6          & 32.8                          & 78.3          & 74.7          \\
DTA  \cite{lee2019drop} &  \XSolidBrush        & 93.7                          & 82.2                          & 85.6                        & 83.8          & 93.0          & 81.0                          & 90.7          & 82.1                           & 95.1          & 78.1                           & 86.4          & 32.1                          & 81.5          & 78.7             \\ \bottomrule     
\end{tabular}}
\vspace{-5mm}
\end{table}
\begin{table}[htp]
    \centering
    \footnotesize
    \caption{Accuracies (\%) on Office-Home for partial-set model adaptation. (`:' denotes `transfer to'.)}\label{tab:partial-set-office-home}
    \scalebox{0.7}{
    \begin{tabular}{
    >{\columncolor[HTML]{FFFFFF}}l |
    >{\columncolor[HTML]{FFFFFF}}c |
    >{\columncolor[HTML]{FFFFFF}}c
    >{\columncolor[HTML]{FFFFFF}}c 
    >{\columncolor[HTML]{FFFFFF}}c 
    >{\columncolor[HTML]{FFFFFF}}c 
    >{\columncolor[HTML]{FFFFFF}}c 
    >{\columncolor[HTML]{FFFFFF}}c 
    >{\columncolor[HTML]{FFFFFF}}c 
    >{\columncolor[HTML]{FFFFFF}}c 
    >{\columncolor[HTML]{FFFFFF}}c 
    >{\columncolor[HTML]{FFFFFF}}c 
    >{\columncolor[HTML]{FFFFFF}}c 
    >{\columncolor[HTML]{FFFFFF}}c 
    >{\columncolor[HTML]{FFF2CC}}c 
    >{\columncolor[HTML]{DAE8FC}}c }
    \toprule
    Method  & DABP  & \cellcolor[HTML]{DAE8FC}Ar:Cl & Ar:Pr         & Ar:Re         & \cellcolor[HTML]{DAE8FC}Cl:Ar & \cellcolor[HTML]{DAE8FC}Cl:Pr & Cl:Re         & \cellcolor[HTML]{DAE8FC}Pr:Ar & \cellcolor[HTML]{DAE8FC}Pr:Cl & Pr:Re         & Re:Ar         & \cellcolor[HTML]{DAE8FC}Re:Cl & Re:Pr         & Avg.          & H. Avg.       \\ \midrule
    ResNet-50 \cite{he2016deep} & - & 44.9                          & 70.5          & 80.7          & 57.5                          & 61.3                          & 67.2          & 60.9                          & 40.8                          & 76.0          & 70.9          & 47.6                          & 76.9          & 62.9          & 52.2          \\
    LNL-OT \cite{asano2019self} &  \Checkmark  & 42.7                          & 64.2          & 71.7          & 57.2                          & 58.5                          & 64.5          & 56.7                          & 41.6                          & 67.5          & 64.2          & 45.1                          & 69.0          & 58.6          & 50.3          \\
    LNL-KL \cite{zhang2021unsupervised} &  \Checkmark   & 38.9                          & 53.8          & 60.5          & 49.2                          & 50.5                          & 55.9          & 50.0                          & 38.9                          & 58.0          & 57.0          & 41.7                          & 59.6          & 51.2          & 44.9          \\
    HD-SHOT \cite{liang2021source} &  \Checkmark  & 51.2                          & 76.2          & 85.7          & 68.8                          & 70.6                          & 77.5          & 69.2                          & 49.6                          & 81.4          & 75.9          & 54.1                          & 80.7          & 70.1          & 60.6          \\
    SD-SHOT \cite{liang2021source} &  \Checkmark  & 54.2                          & 81.8          & 88.9          & 74.8                          & 76.5                          & 81.0          & 73.5                          & 50.6                          & 84.2          & 79.8          & 58.4                          & 83.7          & 74.0          & 64.7          \\
    DINE \cite{liang2022dine} &  \Checkmark     & 58.1                          & 83.4          & 89.2          & \textbf{78.0}                          & 80.0                          & 80.6          & 74.2                          & 56.6                          & 85.9          & 80.6          & \textbf{62.9}                          & 84.8          & 76.2          & 68.3          \\ \midrule
    \textbf{BETA} (Ours) &  \Checkmark     & \textbf{61.7}                 & \textbf{88.5} & \textbf{91.6} & 77.7                 & \textbf{80.1}                 & \textbf{86.3} & \textbf{75.2}                 & \textbf{58.4}                 & \textbf{87.0} & \textbf{81.1} & 61.5                & \textbf{86.7} & \textbf{78.0} & \textbf{69.1} \\ \midrule
    BA$^3$US \cite{liang2020balanced} &  \XSolidBrush  &  60.6 & 83.2 & 88.4 & 71.8 & 72.8 & 83.4 & 75.5 & 61.6 & 86.5 & 79.3 & 62.8 & 86.1 & 76.0 & 67.5\\
    MCC \cite{jin2020minimum} &  \XSolidBrush     & 63.1                          & 80.8          & 86.0          & 70.8                          & 72.1                          & 80.1          & 75.0                          & 60.8                          & 85.9          & 78.6          & 65.2                          & 82.8          & 75.1          & 67.8          \\
    JUMBOT \cite{fatras2021unbalanced} &  \XSolidBrush  & 62.7                          & 77.5          & 84.4          & 76.0                          & 73.3                          & 80.5          & 74.7                          & 60.8                          & 85.1          & 80.2          & 66.5                          & 83.9          & 75.5          & 69.0          \\
    \bottomrule
    \end{tabular}
    }
    \vspace{-3mm}
\end{table}

\subsection{Results}
\textbf{Performance comparison}. We show the results on Office-31, Office-Home, and VisDA-17 in Table~\ref{tab:office-31}, \ref{tab:office-home} and \ref{tab:visda-17}, respectively. The proposed BETA achieves the best performances on all benchmarks. On average, our method outperforms the state-of-the-art methods by 1.8\%, 2.4\%, and 7.0\% on Office-31, Office-Home, and VisDA-17, respectively. Since office-31 is simple with several tasks with 90\%+ accuracies, the improvement is marginal. Whereas, for the challenging VisDA-17, the BETA gains a huge improvement of 7.0\%, and it even outperforms many standard UDA methods~\cite{long2017conditional,lee2019drop,xu2019larger}. This shows that training a target-domain model using pseudo labels with the confirmation bias suppressed can be as effective as the domain alignment techniques.

\textbf{Hard transfer tasks with distant domain shift.} Since our method effectively mitigates the confirmation bias, it works more effectively for the hard tasks with extremely noisy pseudo labels from the source-only model. For the hard tasks with lower than 65\% source-only accuracy (\textit{i.e.}, H. Avg.), it is observed that the BETA outperforms the second-best method by 3.5\%, 4.0\%, and 6.7\% on Office-31, Office-Home, and VisDA-17, respectively, which beats the normal UDA methods. This demonstrates that our method can deal with transfer tasks with distant shifts and poor source-only accuracy by alleviating the negative effect of error accumulation.

\textbf{Partial-set tasks.} Apart from closed-set UDA, we also demonstrate the effectiveness of our method for partial-set UDA tasks. To this end, we select the first 25 classes in alphabetical order as the target domain from Office-Home. As shown in Table~\ref{tab:partial-set-office-home}, it is seen that LNL-OT and LNL-KL lead to negative transfer due to the label shift. Compared to existing state-of-the-art methods, the proposed BETA achieves the best accuracy of 78.0\%, and even outperforms some standard UDA methods~\cite{fatras2021unbalanced,jin2020minimum}. The improvements for partial-set tasks are not large, as BETA is not tailored to address the label shift.

\begin{table}[h]
\vspace{-7mm}
\caption{Ablation studies of learning objectives and MTN on Office-Home.}\label{tab:ablation}
\scriptsize
\resizebox{\columnwidth}{!}{
    \begin{tabular}{
    ccccc|cccccc
    >{\columncolor[HTML]{DAE8FC}}c}
    \toprule
    $\mathcal{L}_{dd}$ & $\mathcal{L}_{kd}$ & $\mathcal{L}_{mi}$ & $\mathcal{L}_{adv}$ & MTN          & Ar→Cl & Cl→Ar & Cl→Pr & Pr→Ar & Pr→Cl & Re→Cl & H. Avg. \\ \midrule
                       &                    &                    &                                       &              & 44.1  & 54.5  & 63.3  & 52.8  & 41.2  & 46.7  & 50.4    \\
    $\checkmark$       &                    &                    &                                       &              & 55.5  & 65.4  & 76.5  & 64.4  & 50.7  & 58.1  & 61.8    \\
    & $\checkmark$       &                    &                                       &              & 54.6  & 63.8  & 75.3  & 62.3  & 48.0    & 55.7  & 60.0    \\
    $\checkmark$       &                    &                    &                                       & $\checkmark$ & 56.6  & 65.0    & 76.7  & 64.1  & 51.6  & 60.4  & 62.4    \\
    $\checkmark$       &                    & $\checkmark$       &                                       & $\checkmark$ & 55.4  & 67.6  & 78.4  & 65.4  & 54.5  & 58.5  & 63.3    \\
    $\checkmark$       & $\checkmark$       & $\checkmark$       &                                       & $\checkmark$ & 56.8  & 68.1  & 78.7  & 67.3  & 55.3  & 59.2  & 64.2    \\
    $\checkmark$       & $\checkmark$       & $\checkmark$       & $\checkmark$                          & $\checkmark$ & 57.2  & 68.0  & 78.6  & 67.5  & 56.0  & 58.9  & 64.4   \\
    \bottomrule
    \end{tabular}
    }
    \vspace{-3mm}
\end{table}

\begin{figure}[tbp]
	\centering
	\subfigure[Confirmation bias]{\includegraphics[width=0.43\textwidth, angle=0]{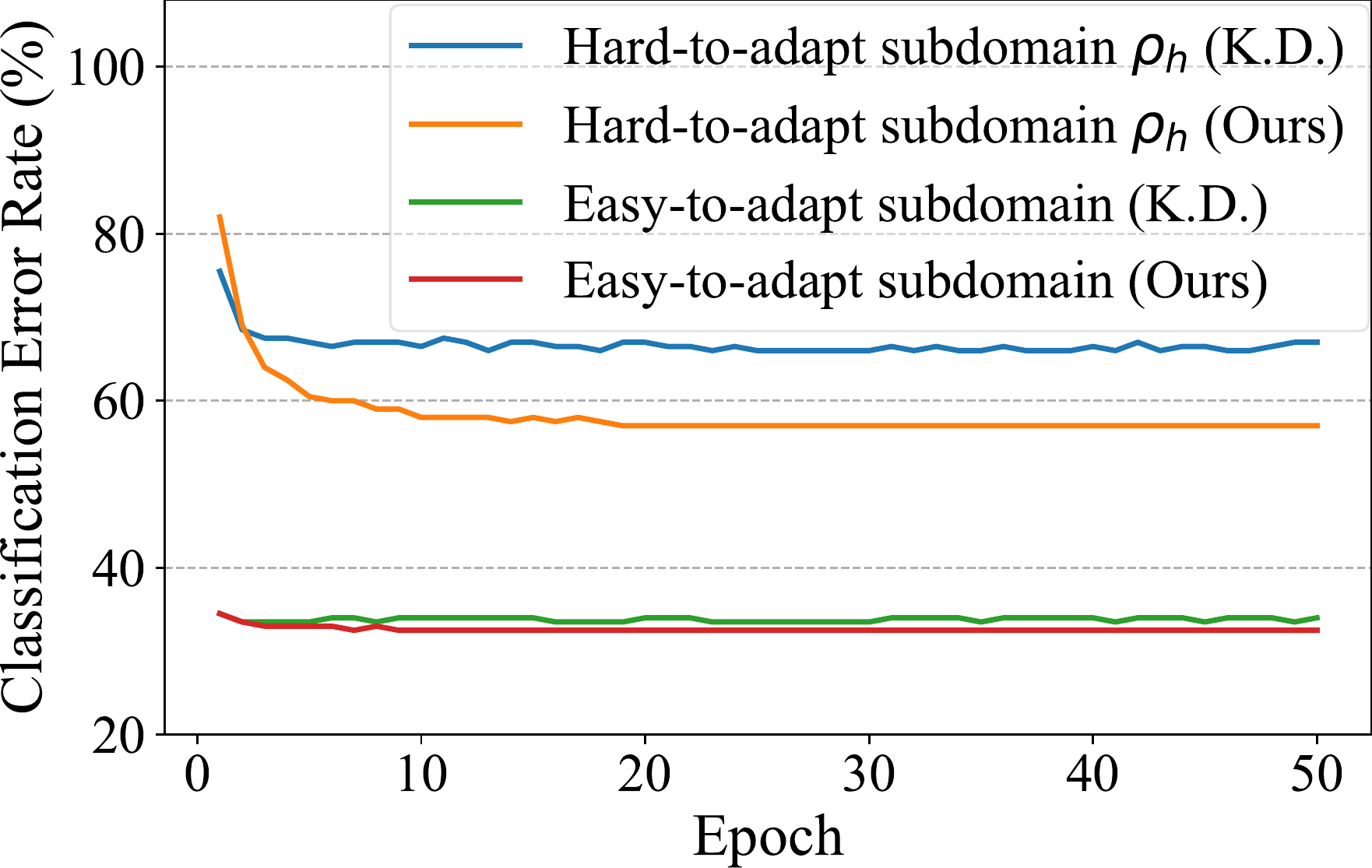}\label{fig:confirmation-bias}}
    \hspace{5mm}
	\subfigure[Studies on $\tau$ and MTNs.]{\includegraphics[width=0.49\textwidth, angle=0]{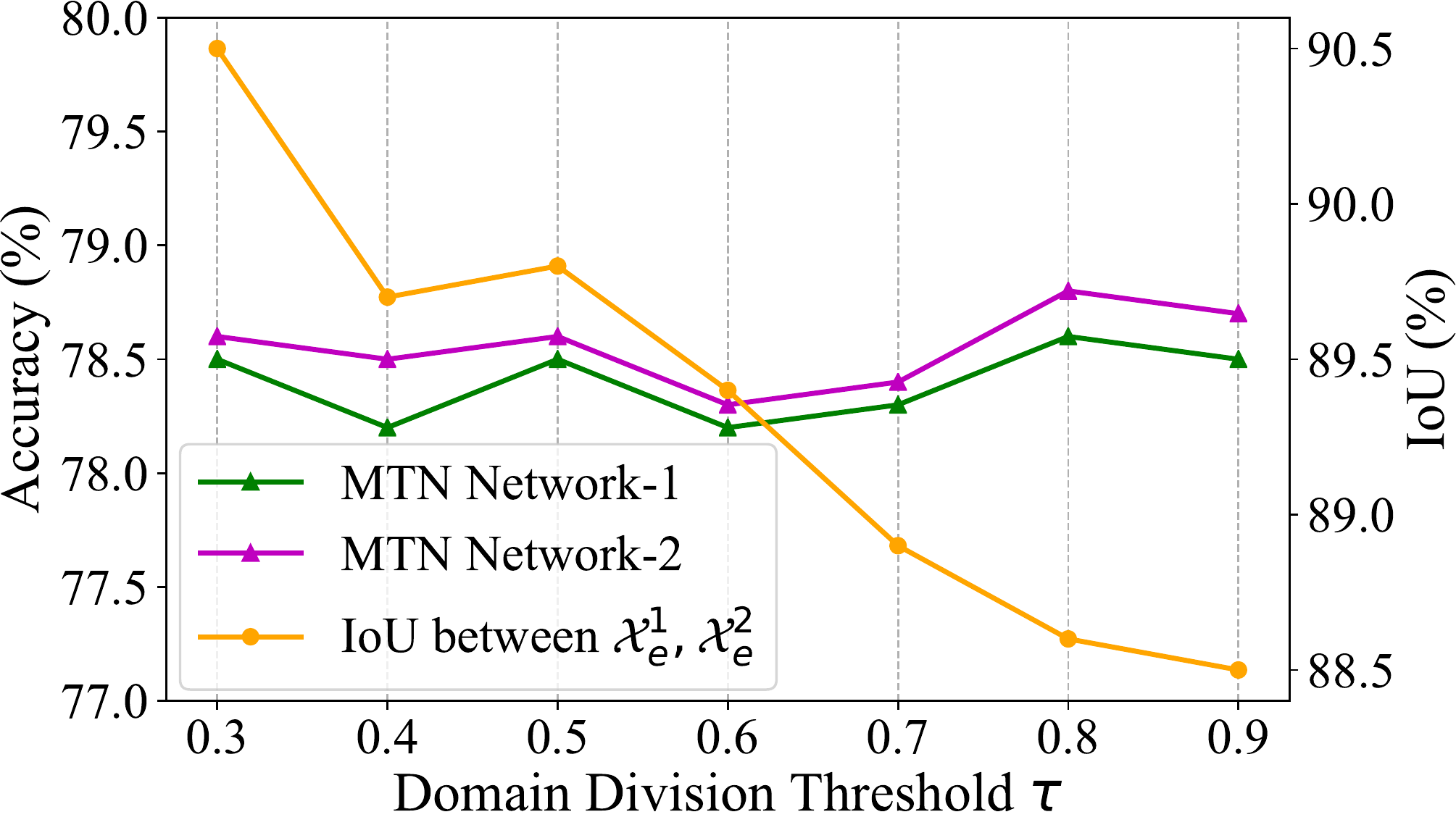}\label{fig:tau}}
	\vspace{-2mm}
	\caption{Quantitative results on the estimated confirmation bias, and hyper-parameter sensitivity.}\label{fig:analytics}
	\vspace{-3mm}
\end{figure}

\begin{figure}[tbp]
	\centering
	{\includegraphics[width=0.31\textwidth, angle=0]{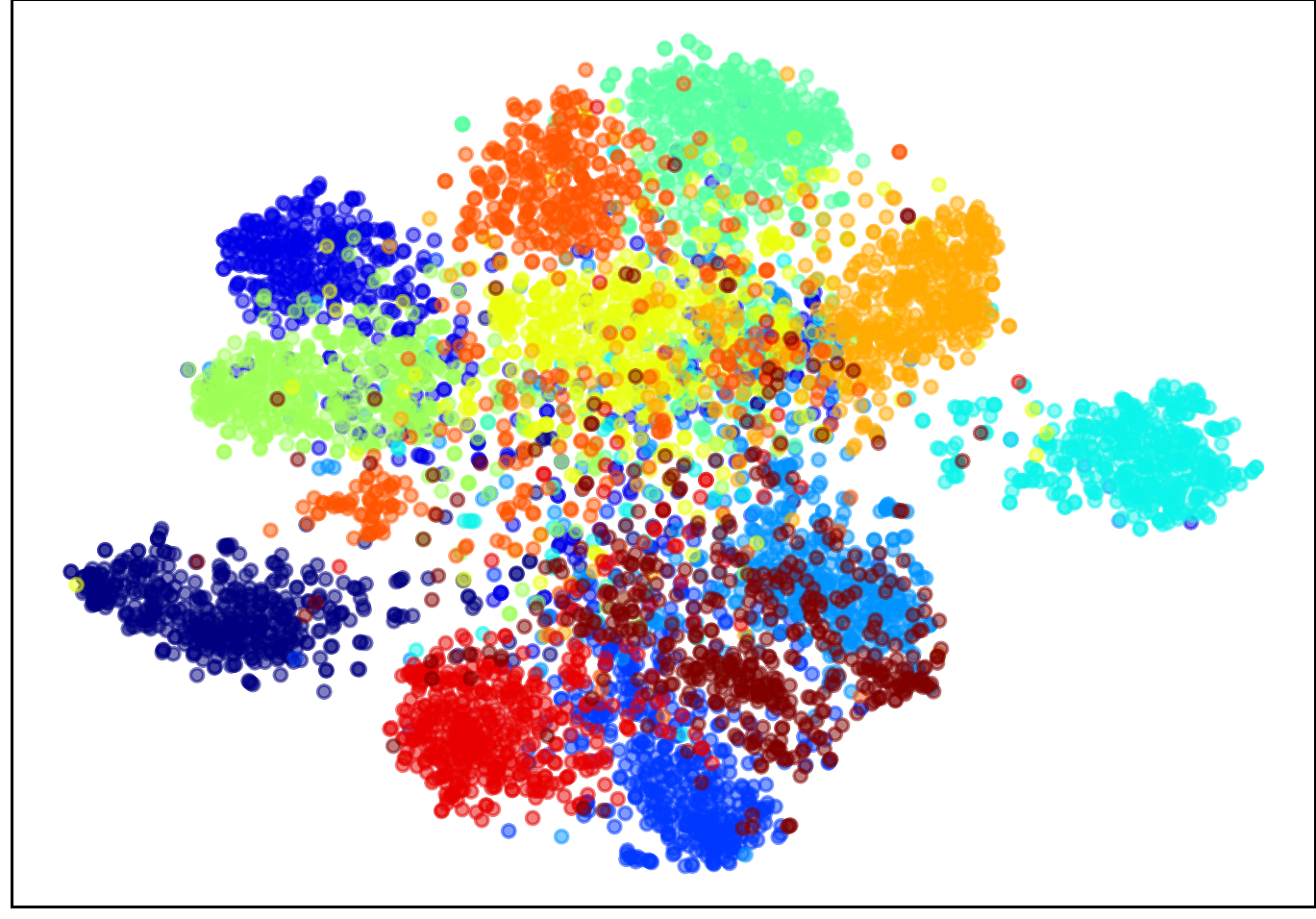}\label{fig:tsne-1}}
	{\includegraphics[width=0.31\textwidth, angle=0]{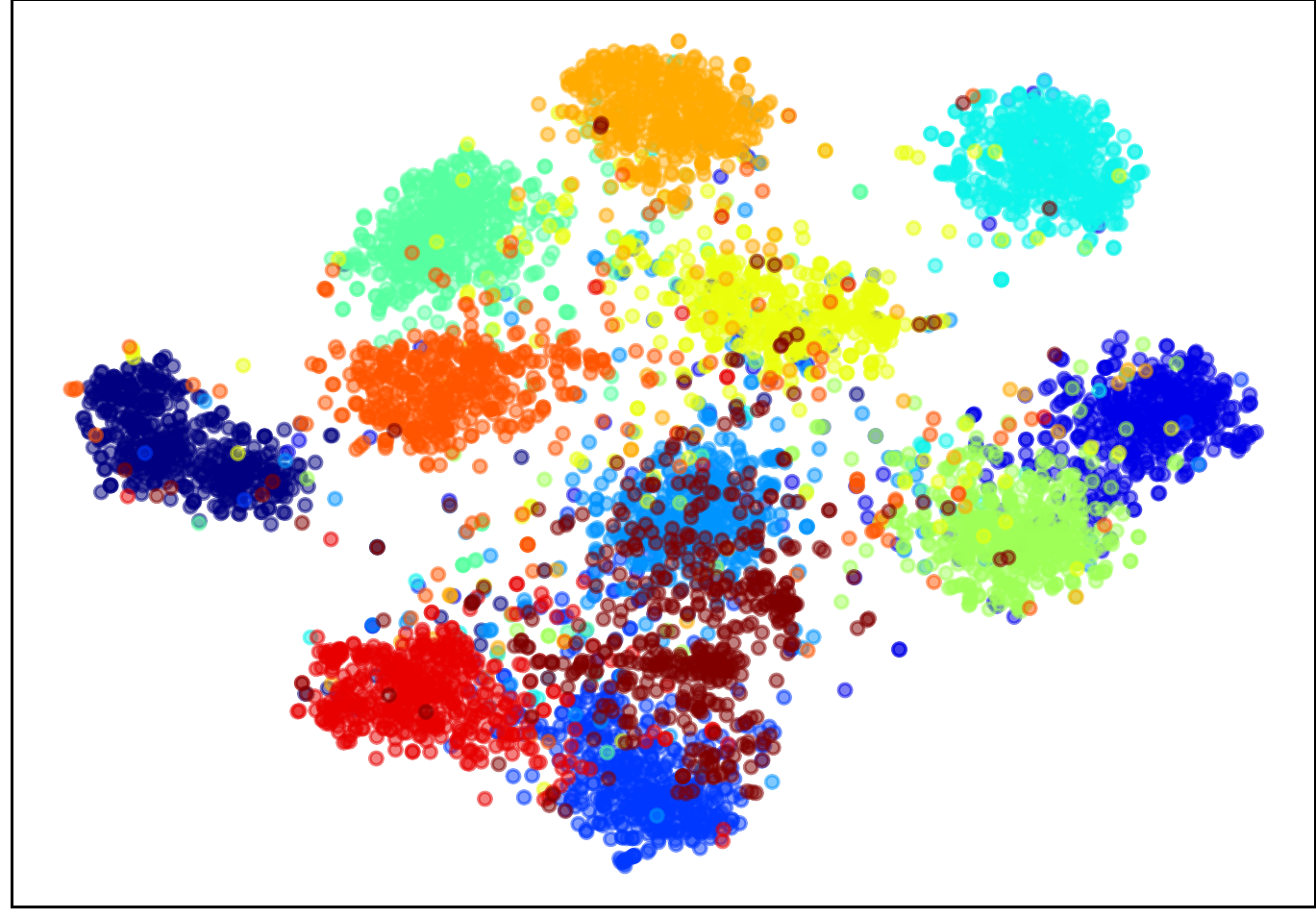}\label{fig:tsne-3}}
	{\includegraphics[width=0.31\textwidth, angle=0]{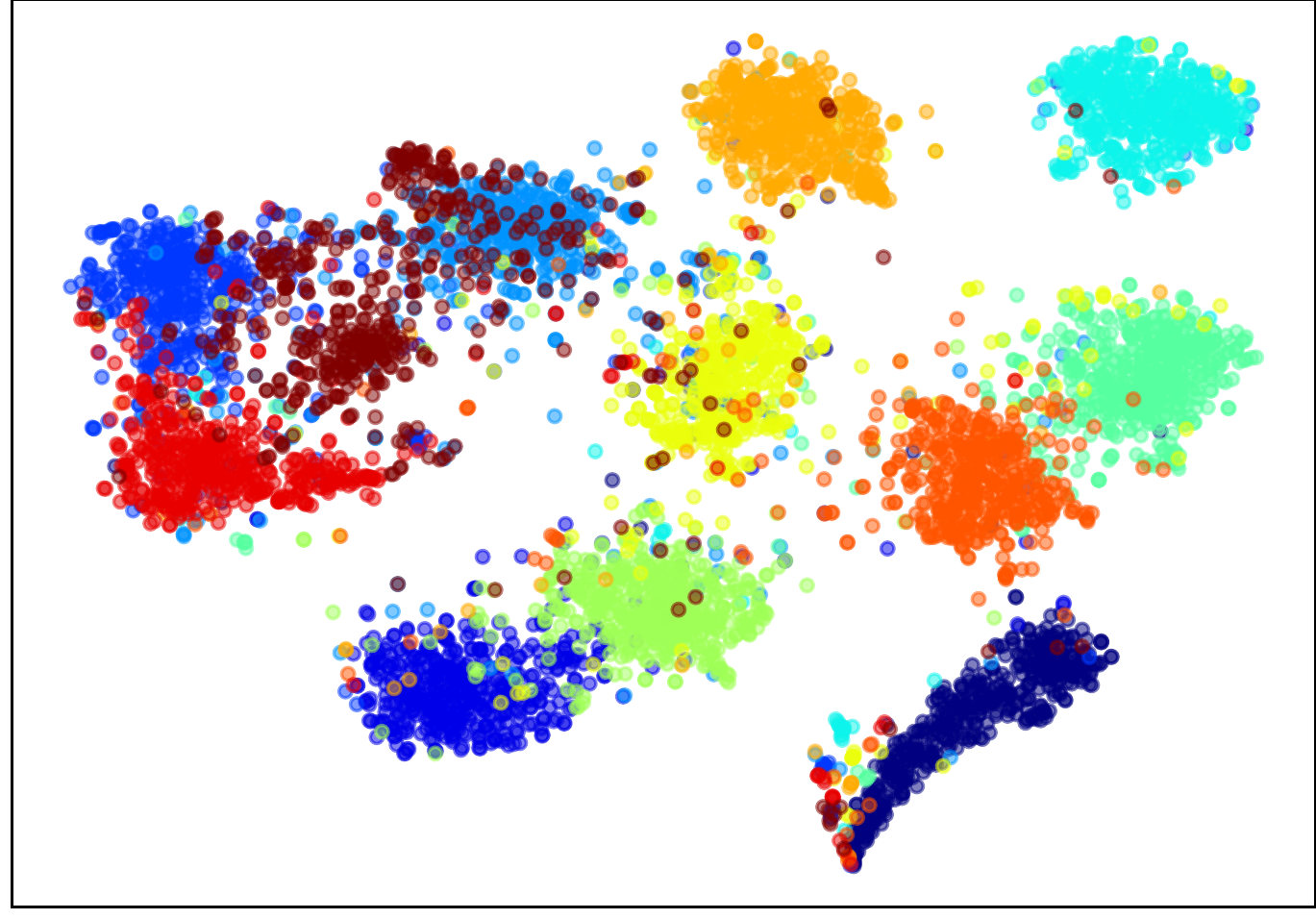}\label{fig:tsne-10}}
	\caption{The t-SNE \cite{tsne2008} visualization of the target domain on the VisDA-17 dataset at the 1st, 3rd, and 10th training epoch (left to right). Each color indicates one category of VisDA-17.}\label{fig:tsne}
	\vspace{-5mm}
\end{figure}

\subsection{Analysis}
\textbf{Ablation study.}
We study the effectiveness of key components in BETA, with results shown in Table~\ref{tab:ablation}. It is seen that the semi-supervised loss enabled by domain division significantly improves the source-only model by 11.4\%. The mutual twin networks, knowledge distillation, and mutual information contribute to 0.6\%, 1.1\%, and 0.9\% improvements, respectively. As the two subdomains drawn from the target domain are quite similar, the distribution discrepancy is not always effective.

\textbf{Confirmation bias.}
We study the confirmation bias using the noise ratio of the two subdomains in terms of the knowledge distillation (K.D.) and BETA on Office-Home (Ar$\to$Cl) to show the effectiveness of the domain division and MTN. As shown in Figure~\ref{fig:confirmation-bias}, the error rate of K.D. only drops for the first a few epochs and then stops decreasing. Whereas, the error rate of BETA keeps decreasing for about 20 epochs since the confirmation bias is iteratively suppressed. The error gap between K.D. and ours on the hard-to-adapt and easy-to-adapt subdomain reaches around 10\% and 3\%, respectively, validating that our method reduces the error rate $\rho_h$ and minimizes the target error in Theorem 1. In Figure~\ref{fig:tsne}, it is observed that the clusters get tighter with clearer boundaries, though there still exists some intrinsic confusion among some classes that remains to be tackled in the future.

\textbf{Hyper-parameter sensitivity and MTN .}
We study the hyper-parameter $\tau$ on Office-Home (Cl$\to$Pr) across three runs. We choose $\tau$ ranging from 0.3 to 0.9, as too small $\tau$ leads to noisy domain division while very large $\tau$ leads to a very small number of samples at the easy-to-adapt subdomain. As shown in Figure~\ref{fig:tau}, the accuracies of BETA range from 78.2\% to 78.8\%, and the best result is achieved at 0.8. For the MTN module, it is observed that the two networks of BETA achieve similar trends over different $\tau$, and one network slightly outperforms another consistently. We further plot the Intersection over Union (IoU) between two easy-to-adapt clean subdomains $\mathcal{X}_e^1,\mathcal{X}_e^2$ generated by domain division, and it decreases as $\tau$ gets greater, which means that a larger $\tau$ leads to more difference of the domain division. The diverged domain division can better mitigate the error accumulation for MTN. Thus, the best result at $\tau=0.8$ is a reasonable trade-off between the divergence of two domain divisions and the sample number of the easy-to-adapt clean subdomain. 


\vspace{-2mm}
\section{Conclusion}
\vspace{-2mm}
In this work, we propose BETA which learns the target domain as a partially-noisy semi-supervised learning task for black-box model adaptation. By conciliating the noisy label learning with knowledge distillation, we divide the target domain into two subdomains and leverage semi-supervised learning improved with subdomain refinement to mitigate the confirmation bias. The synergistic twin networks are designed to increase the diversity and rectify wrong predictions. Besides, we derive a theorem to demonstrate the shrinking error bound of the target domain. Extensive experiments over different backbones and learning setups show that BETA effectively suppresses the noise accumulation and brings significant improvement, achieving state-of-the-art performance on all benchmarks.

\textbf{Acknowledge.} This research is supported by the Presidential Postdoctoral Fellowship in Nanyang Technological University, Singapore. This research is jointly supported by the National Research Foundation, Singapore under its AI Singapore Programme (AISG Award No: AISG2-PhD-2021-08-008) and T1 251RES2114. We thank Google TFRC for supporting us to get access to the Cloud TPUs. We thank CSCS (Swiss National Supercomputing Centre) for supporting us to get access to the Piz Daint supercomputer. We thank TACC (Texas Advanced Computing Center) for supporting us to get access to the Longhorn supercomputer and the Frontera supercomputer. We thank LuxProvide (Luxembourg national supercomputer HPC organization) for supporting us to get access to the MeluXina supercomputer.

{\small
\bibliographystyle{splncs04}
\bibliography{references}}

\begin{thebibliography}{10}
\providecommand{\url}[1]{\texttt{#1}}
\providecommand{\urlprefix}{URL }
\providecommand{\doi}[1]{https://doi.org/#1}

\bibitem{arazo2019unsupervised}
Arazo, E., Ortego, D., Albert, P., O’Connor, N., McGuinness, K.: Unsupervised
  label noise modeling and loss correction. In: International conference on
  machine learning. pp. 312--321. PMLR (2019)

\bibitem{arazo2020pseudo}
Arazo, E., Ortego, D., Albert, P., O’Connor, N.E., McGuinness, K.:
  Pseudo-labeling and confirmation bias in deep semi-supervised learning. In:
  2020 International Joint Conference on Neural Networks (IJCNN). pp.~1--8.
  IEEE (2020)

\bibitem{arpit2017closer}
Arpit, D., Jastrz{\k{e}}bski, S., Ballas, N., Krueger, D., Bengio, E., Kanwal,
  M.S., Maharaj, T., Fischer, A., Courville, A., Bengio, Y., et~al.: A closer
  look at memorization in deep networks. In: International conference on
  machine learning. pp. 233--242. PMLR (2017)

\bibitem{asano2019self}
Asano, Y.M., Rupprecht, C., Vedaldi, A.: Self-labelling via simultaneous
  clustering and representation learning. In: ICLR (2019)

\bibitem{ben2010theory}
Ben-David, S., Blitzer, J., Crammer, K., Kulesza, A., Pereira, F., Vaughan,
  J.W.: A theory of learning from different domains. Machine learning
  \textbf{79}(1-2),  151--175 (2010)

\bibitem{ben2007analysis}
Ben-David, S., Blitzer, J., Crammer, K., Pereira, F.: Analysis of
  representations for domain adaptation. In: Advances in Neural Information
  Processing Systems. pp. 137--144 (2007)

\bibitem{berthelot2019mixmatch}
Berthelot, D., Carlini, N., Goodfellow, I., Papernot, N., Oliver, A., Raffel,
  C.A.: Mixmatch: A holistic approach to semi-supervised learning. Advances in
  Neural Information Processing Systems  \textbf{32} (2019)

\bibitem{chang2019domain}
Chang, W.G., You, T., Seo, S., Kwak, S., Han, B.: Domain-specific batch
  normalization for unsupervised domain adaptation. In: Proceedings of the
  IEEE/CVF conference on Computer Vision and Pattern Recognition. pp.
  7354--7362 (2019)

\bibitem{chen2019transferability}
Chen, X., Wang, S., Long, M., Wang, J.: Transferability vs. discriminability:
  Batch spectral penalization for adversarial domain adaptation. In:
  International Conference on Machine Learning. pp. 1081--1090 (2019)

\bibitem{chen2020self}
Chen, Y., Wei, C., Kumar, A., Ma, T.: Self-training avoids using spurious
  features under domain shift. Advances in Neural Information Processing
  Systems  \textbf{33},  21061--21071 (2020)

\bibitem{crammer2008learning}
Crammer, K., Kearns, M., Wortman, J.: Learning from multiple sources. Journal
  of Machine Learning Research  \textbf{9}(8) (2008)

\bibitem{cubuk2019autoaugment}
Cubuk, E.D., Zoph, B., Mane, D., Vasudevan, V., Le, Q.V.: Autoaugment: Learning
  augmentation strategies from data. In: Proceedings of the IEEE/CVF Conference
  on Computer Vision and Pattern Recognition. pp. 113--123 (2019)

\bibitem{cubuk2020randaugment}
Cubuk, E.D., Zoph, B., Shlens, J., Le, Q.V.: Randaugment: Practical automated
  data augmentation with a reduced search space. In: Proceedings of the
  IEEE/CVF Conference on Computer Vision and Pattern Recognition Workshops. pp.
  702--703 (2020)

\bibitem{cui2020heuristic}
Cui, S., Jin, X., Wang, S., He, Y., Huang, Q.: Heuristic domain adaptation.
  Advances in Neural Information Processing Systems  \textbf{33},  7571--7583
  (2020)

\bibitem{cui2020towards}
Cui, S., Wang, S., Zhuo, J., Li, L., Huang, Q., Tian, Q.: Towards
  discriminability and diversity: Batch nuclear-norm maximization under label
  insufficient situations. In: Proceedings of the IEEE/CVF Conference on
  Computer Vision and Pattern Recognition. pp. 3941--3950 (2020)

\bibitem{deng2019cluster}
Deng, Z., Luo, Y., Zhu, J.: Cluster alignment with a teacher for unsupervised
  domain adaptation. In: Proceedings of the IEEE/CVF International Conference
  on Computer Vision. pp. 9944--9953 (2019)

\bibitem{dong2021confident}
Dong, J., Fang, Z., Liu, A., Sun, G., Liu, T.: Confident anchor-induced
  multi-source free domain adaptation. Advances in Neural Information
  Processing Systems  \textbf{34} (2021)

\bibitem{fatras2021unbalanced}
Fatras, K., S{\'e}journ{\'e}, T., Flamary, R., Courty, N.: Unbalanced minibatch
  optimal transport; applications to domain adaptation. In: International
  Conference on Machine Learning. pp. 3186--3197. PMLR (2021)

\bibitem{ganin15}
Ganin, Y., Lempitsky, V.: Unsupervised domain adaptation by backpropagation.
  In: Bach, F., Blei, D. (eds.) Proceedings of the 32nd International
  Conference on Machine Learning. Proceedings of Machine Learning Research,
  vol.~37, pp. 1180--1189. PMLR, Lille, France (07--09 Jul 2015)

\bibitem{ganin2016domain}
Ganin, Y., Ustinova, E., Ajakan, H., Germain, P., Larochelle, H., Laviolette,
  F., Marchand, M., Lempitsky, V.: Domain-adversarial training of neural
  networks. The Journal of Machine Learning Research  \textbf{17}(1),
  2096--2030 (2016)

\bibitem{grandvalet2005semi}
Grandvalet, Y., Bengio, Y.: Semi-supervised learning by entropy minimization.
  In: Advances in neural information processing systems. pp. 529--536 (2005)

\bibitem{gu2020spherical}
Gu, X., Sun, J., Xu, Z.: Spherical space domain adaptation with robust
  pseudo-label loss. In: Proceedings of the IEEE/CVF Conference on Computer
  Vision and Pattern Recognition. pp. 9101--9110 (2020)

\bibitem{guan2021domain}
Guan, H., Liu, M.: Domain adaptation for medical image analysis: a survey. IEEE
  Transactions on Biomedical Engineering  (2021)

\bibitem{he2016deep}
He, K., Zhang, X., Ren, S., Sun, J.: Deep residual learning for image
  recognition. In: Proceedings of the IEEE conference on computer vision and
  pattern recognition. pp. 770--778 (2016)

\bibitem{hinton2015distilling}
Hinton, G., Vinyals, O., Dean, J., et~al.: Distilling the knowledge in a neural
  network. arXiv preprint arXiv:1503.02531  \textbf{2}(7) (2015)

\bibitem{hoffman2018cycada}
Hoffman, J., Tzeng, E., Park, T., Zhu, J.Y., Isola, P., Saenko, K., Efros, A.,
  Darrell, T.: Cycada: Cycle-consistent adversarial domain adaptation. In:
  International conference on machine learning. pp. 1989--1998. PMLR (2018)

\bibitem{huang2021model}
Huang, J., Guan, D., Xiao, A., Lu, S.: Model adaptation: Historical contrastive
  learning for unsupervised domain adaptation without source data. Advances in
  Neural Information Processing Systems  \textbf{34} (2021)

\bibitem{huang2006correcting}
Huang, J., Gretton, A., Borgwardt, K., Sch{\"o}lkopf, B., Smola, A.: Correcting
  sample selection bias by unlabeled data. Advances in neural information
  processing systems  \textbf{19} (2006)

\bibitem{jin2020minimum}
Jin, Y., Wang, X., Long, M., Wang, J.: Minimum class confusion for versatile
  domain adaptation. In: European Conference on Computer Vision. pp. 464--480.
  Springer (2020)

\bibitem{kang2019contrastive}
Kang, G., Jiang, L., Yang, Y., Hauptmann, A.G.: Contrastive adaptation network
  for unsupervised domain adaptation. In: Proceedings of the IEEE/CVF
  Conference on Computer Vision and Pattern Recognition. pp. 4893--4902 (2019)

\bibitem{kim2021fine}
Kim, T., Ko, J., Choi, J., Yun, S.Y., et~al.: Fine samples for learning with
  noisy labels. Advances in Neural Information Processing Systems  \textbf{34}
  (2021)

\bibitem{koniusz2017domain}
Koniusz, P., Tas, Y., Porikli, F.: Domain adaptation by mixture of alignments
  of second-or higher-order scatter tensors. In: Proceedings of the IEEE
  conference on computer vision and pattern recognition. pp. 4478--4487 (2017)

\bibitem{kundu2019adapt}
Kundu, J.N., Lakkakula, N., Babu, R.V.: Um-adapt: Unsupervised multi-task
  adaptation using adversarial cross-task distillation. In: Proceedings of the
  IEEE/CVF International Conference on Computer Vision. pp. 1436--1445 (2019)

\bibitem{kuzborskij2013stability}
Kuzborskij, I., Orabona, F.: Stability and hypothesis transfer learning. In:
  International Conference on Machine Learning. pp. 942--950. PMLR (2013)

\bibitem{lee2019drop}
Lee, S., Kim, D., Kim, N., Jeong, S.G.: Drop to adapt: Learning discriminative
  features for unsupervised domain adaptation. In: Proceedings of the IEEE/CVF
  International Conference on Computer Vision. pp. 91--100 (2019)

\bibitem{li2019dividemix}
Li, J., Socher, R., Hoi, S.C.: Dividemix: Learning with noisy labels as
  semi-supervised learning. In: International Conference on Learning
  Representations (2019)

\bibitem{li2020model}
Li, R., Jiao, Q., Cao, W., Wong, H.S., Wu, S.: Model adaptation: Unsupervised
  domain adaptation without source data. In: Proceedings of the IEEE/CVF
  Conference on Computer Vision and Pattern Recognition. pp. 9641--9650 (2020)

\bibitem{liang2018aggregating}
Liang, J., He, R., Sun, Z., Tan, T.: Aggregating randomized
  clustering-promoting invariant projections for domain adaptation. IEEE
  transactions on pattern analysis and machine intelligence  \textbf{41}(5),
  1027--1042 (2018)

\bibitem{liang2020we}
Liang, J., Hu, D., Feng, J.: Do we really need to access the source data?
  source hypothesis transfer for unsupervised domain adaptation. In:
  International Conference on Machine Learning. pp. 6028--6039. PMLR (2020)

\bibitem{liang2021domain}
Liang, J., Hu, D., Feng, J.: Domain adaptation with auxiliary target
  domain-oriented classifier. In: Proceedings of the IEEE/CVF Conference on
  Computer Vision and Pattern Recognition. pp. 16632--16642 (2021)

\bibitem{liang2022dine}
Liang, J., Hu, D., Feng, J., He, R.: Dine: Domain adaptation from single and
  multiple black-box predictors. In: Proceedings of the IEEE/CVF Conference on
  Computer Vision and Pattern Recognition (2022)

\bibitem{liang2021source}
Liang, J., Hu, D., Wang, Y., He, R., Feng, J.: Source data-absent unsupervised
  domain adaptation through hypothesis transfer and labeling transfer. IEEE
  Transactions on Pattern Analysis and Machine Intelligence  (2021)

\bibitem{liang2020balanced}
Liang, J., Wang, Y., Hu, D., He, R., Feng, J.: A balanced and uncertainty-aware
  approach for partial domain adaptation. In: European Conference on Computer
  Vision. pp. 123--140. Springer (2020)

\bibitem{liu2020learning}
Liu, H., Long, M., Wang, J., Wang, Y.: Learning to adapt to evolving domains.
  Advances in Neural Information Processing Systems  \textbf{33},  22338--22348
  (2020)

\bibitem{liu2021cycle}
Liu, H., Wang, J., Long, M.: Cycle self-training for domain adaptation.
  Advances in Neural Information Processing Systems  \textbf{34} (2021)

\bibitem{long2017conditional}
Long, M., Cao, Z., Wang, J., Jordan, M.I.: Conditional adversarial domain
  adaptation. In: Advances in Neural Information Processing Systems (2018)

\bibitem{long2016deep}
Long, M., Zhu, H., Wang, J., Jordan, M.I.: Deep transfer learning with joint
  adaptation networks. In: International Conference on Machine Learning (ICML)
  (2017)

\bibitem{luo2021conditional}
Luo, Y.W., Ren, C.X.: Conditional bures metric for domain adaptation. In:
  Proceedings of the IEEE/CVF Conference on Computer Vision and Pattern
  Recognition. pp. 13989--13998 (2021)

\bibitem{tsne2008}
Van~der Maaten, L., Hinton, G.: Visualizing data using t-sne. Journal of
  machine learning research  \textbf{9}(11) (2008)

\bibitem{morerio2020generative}
Morerio, P., Volpi, R., Ragonesi, R., Murino, V.: Generative pseudo-label
  refinement for unsupervised domain adaptation. In: Proceedings of the
  IEEE/CVF Winter Conference on Applications of Computer Vision. pp. 3130--3139
  (2020)

\bibitem{pan2010domain}
Pan, S.J., Tsang, I.W., Kwok, J.T., Yang, Q.: Domain adaptation via transfer
  component analysis. IEEE transactions on neural networks  \textbf{22}(2),
  199--210 (2010)

\bibitem{pan2010survey}
Pan, S.J., Yang, Q., et~al.: A survey on transfer learning. IEEE Transactions
  on knowledge and data engineering  \textbf{22}(10),  1345--1359 (2010)

\bibitem{paszke2019pytorch}
Paszke, A., Gross, S., Massa, F., Lerer, A., Bradbury, J., Chanan, G., Killeen,
  T., Lin, Z., Gimelshein, N., Antiga, L., et~al.: Pytorch: An imperative
  style, high-performance deep learning library. Advances in neural information
  processing systems  \textbf{32} (2019)

\bibitem{peng2017visda}
Peng, X., Usman, B., Kaushik, N., Hoffman, J., Wang, D., Saenko, K.: Visda: The
  visual domain adaptation challenge. arXiv preprint arXiv:1710.06924  (2017)

\bibitem{qiao2018deep}
Qiao, S., Shen, W., Zhang, Z., Wang, B., Yuille, A.: Deep co-training for
  semi-supervised image recognition. In: Proceedings of the european conference
  on computer vision (eccv). pp. 135--152 (2018)

\bibitem{saenko2010adapting}
Saenko, K., Kulis, B., Fritz, M., Darrell, T.: Adapting visual category models
  to new domains. In: European conference on computer vision. pp. 213--226.
  Springer (2010)

\bibitem{saito2020universal}
Saito, K., Kim, D., Sclaroff, S., Saenko, K.: Universal domain adaptation
  through self supervision. Advances in neural information processing systems
  \textbf{33},  16282--16292 (2020)

\bibitem{saito2017asymmetric}
Saito, K., Ushiku, Y., Harada, T.: Asymmetric tri-training for unsupervised
  domain adaptation. In: International Conference on Machine Learning. pp.
  2988--2997. PMLR (2017)

\bibitem{saito2018maximum}
Saito, K., Watanabe, K., Ushiku, Y., Harada, T.: Maximum classifier discrepancy
  for unsupervised domain adaptation. In: Proceedings of the IEEE Conference on
  Computer Vision and Pattern Recognition. pp. 3723--3732 (2018)

\bibitem{shu2018dirt}
Shu, R., Bui, H.H., Narui, H., Ermon, S.: A dirt-t approach to unsupervised
  domain adaptation. In: Proc. 6th International Conference on Learning
  Representations (2018)

\bibitem{sohn2020fixmatch}
Sohn, K., Berthelot, D., Carlini, N., Zhang, Z., Zhang, H., Raffel, C.A.,
  Cubuk, E.D., Kurakin, A., Li, C.L.: Fixmatch: Simplifying semi-supervised
  learning with consistency and confidence. Advances in Neural Information
  Processing Systems  \textbf{33},  596--608 (2020)

\bibitem{song2022learning}
Song, H., Kim, M., Park, D., Shin, Y., Lee, J.G.: Learning from noisy labels
  with deep neural networks: A survey. IEEE Transactions on Neural Networks and
  Learning Systems  (2022)

\bibitem{sugiyama2007direct}
Sugiyama, M., Nakajima, S., Kashima, H., Buenau, P., Kawanabe, M.: Direct
  importance estimation with model selection and its application to covariate
  shift adaptation. Advances in neural information processing systems
  \textbf{20} (2007)

\bibitem{sun2016deep}
Sun, B., Saenko, K.: Deep coral: Correlation alignment for deep domain
  adaptation. In: European Conference on Computer Vision. pp. 443--450.
  Springer (2016)

\bibitem{tarvainen2017mean}
Tarvainen, A., Valpola, H.: Mean teachers are better role models:
  Weight-averaged consistency targets improve semi-supervised deep learning
  results. Advances in neural information processing systems  \textbf{30}
  (2017)

\bibitem{tzeng2015simultaneous}
Tzeng, E., Hoffman, J., Darrell, T., Saenko, K.: Simultaneous deep transfer
  across domains and tasks. In: Proceedings of the IEEE International
  Conference on Computer Vision. pp. 4068--4076 (2015)

\bibitem{tzeng2017adversarial}
Tzeng, E., Hoffman, J., Saenko, K., Darrell, T.: Adversarial discriminative
  domain adaptation. In: Computer Vision and Pattern Recognition (CVPR).
  vol.~1, p.~4 (2017)

\bibitem{TzengHZSD14}
Tzeng, E., Hoffman, J., Zhang, N., Saenko, K., Darrell, T.: Deep domain
  confusion: Maximizing for domain invariance. CoRR  \textbf{abs/1412.3474}
  (2014), \url{http://arxiv.org/abs/1412.3474}

\bibitem{venkateswara2017deep}
Venkateswara, H., Eusebio, J., Chakraborty, S., Panchanathan, S.: Deep hashing
  network for unsupervised domain adaptation. In: Proc. CVPR. pp. 5018--5027
  (2017)

\bibitem{wang2022reliable}
Wang, K., Peng, X., Yang, S., Yang, J., Zhu, Z., Wang, X., You, Y.: Reliable
  label correction is a good booster when learning with extremely noisy labels.
  arXiv preprint arXiv:2205.00186  (2022)

\bibitem{wang2019transferable}
Wang, X., Jin, Y., Long, M., Wang, J., Jordan, M.I.: Transferable
  normalization: Towards improving transferability of deep neural networks.
  Advances in neural information processing systems  \textbf{32} (2019)

\bibitem{xie2018learning}
Xie, S., Zheng, Z., Chen, L., Chen, C.: Learning semantic representations for
  unsupervised domain adaptation. In: International conference on machine
  learning. pp. 5423--5432. PMLR (2018)

\bibitem{xu2019larger}
Xu, R., Li, G., Yang, J., Lin, L.: Larger norm more transferable: An adaptive
  feature norm approach for unsupervised domain adaptation. In: Proceedings of
  the IEEE/CVF International Conference on Computer Vision. pp. 1426--1435
  (2019)

\bibitem{xu2021partial}
Xu, Y., Yang, J., Cao, H., Chen, Z., Li, Q., Mao, K.: Partial video domain
  adaptation with partial adversarial temporal attentive network. In:
  Proceedings of the IEEE/CVF International Conference on Computer Vision. pp.
  9332--9341 (2021)

\bibitem{yang2021advancing}
Yang, J., Yang, J., Wang, S., Cao, S., Zou, H., Xie, L.: Advancing imbalanced
  domain adaptation: Cluster-level discrepancy minimization with a
  comprehensive benchmark. IEEE Transactions on Cybernetics  (2021)

\bibitem{yang2021robust}
Yang, J., Zou, H., Zhou, Y., Xie, L.: Robust adversarial discriminative domain
  adaptation for real-world cross-domain visual recognition. Neurocomputing
  \textbf{433},  28--36 (2021)

\bibitem{yang2020mind}
Yang, J., Zou, H., Zhou, Y., Zeng, Z., Xie, L.: Mind the discriminability:
  Asymmetric adversarial domain adaptation. In: European Conference on Computer
  Vision. pp. 589--606. Springer (2020)

\bibitem{zhang2021unsupervised}
Zhang, H., Zhang, Y., Jia, K., Zhang, L.: Unsupervised domain adaptation of
  black-box source models. arXiv preprint arXiv:2101.02839  (2021)

\bibitem{zhang2018mixup}
Zhang, H., Cisse, M., Dauphin, Y.N., Lopez-Paz, D.: mixup: Beyond empirical
  risk minimization. In: International Conference on Learning Representations
  (2018)

\bibitem{zhang2017curriculum}
Zhang, Y., David, P., Gong, B.: Curriculum domain adaptation for semantic
  segmentation of urban scenes. In: Proceedings of the IEEE international
  conference on computer vision. pp. 2020--2030 (2017)

\bibitem{zhang2020collaborative}
Zhang, Y., Wei, Y., Wu, Q., Zhao, P., Niu, S., Huang, J., Tan, M.:
  Collaborative unsupervised domain adaptation for medical image diagnosis.
  IEEE Transactions on Image Processing  \textbf{29},  7834--7844 (2020)

\bibitem{zhang2019bridging}
Zhang, Y., Liu, T., Long, M., Jordan, M.: Bridging theory and algorithm for
  domain adaptation. In: International Conference on Machine Learning. pp.
  7404--7413 (2019)

\bibitem{zhou2020domain}
Zhou, B., Kalra, N., Kr{\"a}henb{\"u}hl, P.: Domain adaptation through task
  distillation. In: European Conference on Computer Vision. pp. 664--680.
  Springer (2020)

\bibitem{zou2021unsupervised}
Zou, H., Yang, J., Wu, X.: Unsupervised energy-based adversarial domain
  adaptation for cross-domain text classification. In: Findings of the
  Association for Computational Linguistics: ACL-IJCNLP 2021. pp. 1208--1218
  (2021)

\bibitem{zou2019consensus}
Zou, H., Zhou, Y., Yang, J., Liu, H., Das, H.P., Spanos, C.J.: Consensus
  adversarial domain adaptation. In: Proceedings of the AAAI Conference on
  Artificial Intelligence. vol.~33, pp. 5997--6004 (2019)

\bibitem{zou2018unsupervised}
Zou, Y., Yu, Z., Kumar, B., Wang, J.: Unsupervised domain adaptation for
  semantic segmentation via class-balanced self-training. In: Proceedings of
  the European conference on computer vision (ECCV). pp. 289--305 (2018)

\end{thebibliography}

\section*{Appendix}
\subsection*{Proof of Theorem 1}

We prove the \textbf{Theorem 1} which extends the learning theories of domain adaptation \cite{ben2010theory} for the black-box domain adaptation and provides theoretical justifications for our method.

Denote $\mathcal{X}_t\sim \mathcal{D}_T$ as the target domain with its sample distribution. $\mathcal{X}_e\sim \mathcal{D}_e$ and $\mathcal{X}_h\sim \mathcal{D}_h$ denote the easy-to-adapt clean subdomain and the hard-to-adapt noisy subdomain with their corresponding sample distributions, respectively. Denote $y_e,y_h$ and $\hat{y}_e,\hat{y}_h$ as the ground truth labels and the pseudo labels of $\mathcal{X}_e, \mathcal{X}_h$, respectively. Let $h$ denote a hypothesis. As our method performs training on a mixture of the clean set and the noisy set with pseudo labels, the error of our method can be formulated as a convex combination of the errors of the clean set and the noisy set:
\begin{equation}
    \epsilon_\alpha(h) = \alpha \epsilon_e(h,\hat{y}_e)+(1-\alpha)\epsilon_h(h,\hat{y}_h),
\end{equation}
where $\alpha$ is the trade-off hyper-parameter, and $\epsilon_e(h,\hat{y}_e),\epsilon_h(h,\hat{y}_h)$ represents the expected error of the easy-to-adapt clean set $\mathcal{X}_e$ and the hard-to-adapt noisy set $\mathcal{X}_h$, respectively, defined by
\begin{align}
    & \epsilon_e(h,\hat{y}_e)=\mathbb{E}_{x\sim \mathcal{D}_e} [|h(x)-\hat{y}_e|]\\
    & \epsilon_h(h,\hat{y}_h)=\mathbb{E}_{x\sim \mathcal{D}_h} [|h(x)-\hat{y}_h|].
\end{align}
We use the shorthand $\epsilon_e(h)=\epsilon_e(h,f_e)$ in the proof.

Then, we derive an upper bound of how the error $\epsilon_\alpha(h)$ is close to an oracle error of the target domain $\epsilon_t(h,y_t)$ where $y_t$ is the ground truth labels of the target domain, which is illustrated in \textbf{Theorem 1}:

{\bf Theorem 1} {\it
Let $h$ be a hypothesis in class $\mathcal{H}$. Then
\begin{equation}
|\epsilon_\alpha(h)-\epsilon_t(h,y_t)|\leq \alpha(d_{\mathcal{H}\triangle\mathcal{H}}(\mathcal{D}_e,\mathcal{D}_h)+\lambda+\hat{\lambda})+\rho_h,
\end{equation}
where the ideal risk is the combined error of the ideal joint hypothesis $\lambda=\epsilon_e(h^*)+\epsilon_h(h^*)$, the distribution discrepancy $d_\mathcal{\mathcal{H}\triangle\mathcal{H}}(\mathcal{D}_e,\mathcal{D}_h)=2\sup_{h,h' \in \mathcal{H}} |\mathbb{E}_{x\sim \mathcal{D}_e}[h(x)\ne h'(x)] - \mathbb{E}_{x\sim \mathcal{D}_h}[h(x)\ne h'(x)]|$, and $\rho_h$ denote the pseudo label rate of $\hat{y}_h$. The ideal joint hypothesis is given by $h^*=\arg \min_{h\in\mathcal{H}}(\epsilon_e(h)+\epsilon_h(h))$, deriving the ideal risk $\lambda=\epsilon_e(h^*)+\epsilon_h(h^*)$ and the pseudo risk $\hat{\lambda}=\epsilon_e(h^*,\hat{y}_e)+\epsilon_h(h^*,\hat{y}_h)$.
}

\textit{Proof:}
\begin{align}\nonumber
    & |\epsilon_\alpha(h)-\epsilon_t(h,y_t)| \\
    & = |\alpha \epsilon_e(h,\hat{y}_e)+(1-\alpha) \epsilon_h(h,\hat{y}_h)-\alpha \epsilon_e(h,y_e)-(1-\alpha)\epsilon_h(h,y_h)|\\
    & \le \alpha(|\epsilon_e(h,y_e)-\epsilon_h(h,y_h)|+|\epsilon_e(h,\hat{y}_e)-\epsilon_h(h,\hat{y}_h)|)+|\epsilon_h(h,\hat{y}_h)-\epsilon_h(h,y_h)|\\
    & = \alpha(\epsilon_a+\epsilon_b) + \epsilon_c
\end{align}

Then we seek the upper bound of $\epsilon_a,\epsilon_b,\epsilon_c$ by applying the triangle inequality for classification errors \cite{crammer2008learning} as stated in \textbf{Lemma 1}.
\begin{lemma}
For any hypotheses $f_1,f_2,f_3$ in class $\mathcal{H}$, 
\begin{equation}
    \epsilon(f_1,f_2) \le \epsilon(f_1,f_3)+\epsilon(f_2,f_3).
\end{equation}
\end{lemma}

For $\epsilon_a$, 
\begin{align}\nonumber
    & \epsilon_a=|\epsilon_e(h,y_e)-\epsilon_h(h,y_h)| \\
    & \le |\epsilon_e(h,y_e)-\epsilon_e(h,h^*)|+|\epsilon_e(h,h^*)-\epsilon_h(h,h^*)|+|\epsilon_h(h,h^*)-\epsilon_h(h,y_h)| \\
    & \le \epsilon_e(h^*)+|\epsilon_e(h,h^*)-\epsilon_h(h,h^*)|+\epsilon_h(h^*) \\
    & \le \frac{1}{2}d_{\mathcal{H}\triangle\mathcal{H}}(\mathcal{D}_e,\mathcal{D}_h)+\lambda
\end{align}

For $\epsilon_b$,
\begin{align}\nonumber
    & \epsilon_b=|\epsilon_e(h,\hat{y}_e)-\epsilon_h(h,\hat{y}_h)|\\
    & \le \epsilon_e(h^*,\hat{y}_e)+|\epsilon_e(h,h^*)-\epsilon_h(h,h^*)|+\epsilon_h(h^*,\hat{y}_h) \\
    & \le \frac{1}{2}d_{\mathcal{H}\triangle\mathcal{H}}(\mathcal{D}_c,\mathcal{D}_n)+(\epsilon_e(h^*,\hat{y}_e)+\epsilon_h(h^*,\hat{y}_h)) \\
    & \le \frac{1}{2}d_{\mathcal{H}\triangle\mathcal{H}}(\mathcal{D}_e,\mathcal{D}_h)+\hat{\lambda} \\
\end{align}

For $\epsilon_c$,
\begin{align}\nonumber
    & \epsilon_c= |\epsilon_h(h,\hat{y}_h)-\epsilon_h(h,y_h)| \le |\epsilon_h(\hat{y}_h,y_h)| = \rho_h
\end{align}

By summarizing $\epsilon_a,\epsilon_b,\epsilon_c$, we yield the inequality in \textbf{Theorem 1}:
\begin{align}
    & |\epsilon_\alpha(h)-\epsilon_t(h,y_t)| \\
    & \le \alpha[(\frac{1}{2}d_{\mathcal{H}\triangle\mathcal{H}}(\mathcal{D}_e,\mathcal{D}_h)+\lambda)+(\frac{1}{2}d_{\mathcal{H}\triangle\mathcal{H}}(\mathcal{D}_e,\mathcal{D}_h) + \hat{\lambda})]+\rho_h \\
    & = \alpha(d_{\mathcal{H}\triangle\mathcal{H}}(\mathcal{D}_e,\mathcal{D}_h)+\lambda+\hat{\lambda})+\rho_h
\end{align}
$\square$

Furthermore, the pseudo risk is bounded by the ideal risk, the pseudo rate of the clean set $\rho_e$ and the noisy set $\rho_h$, derived as follows:
\begin{align}
    & \hat{\lambda}=\epsilon_e(h^*,\hat{y}_e)+\epsilon_h(h^*,\hat{y}_h) \\
    & \le (\epsilon_e(h^*,y_e)+\epsilon_e(y_e,\hat{y}_e))+(\epsilon_h(h^*,y_h)+\epsilon_h(y_h,\hat{y}_h)) \\
    & = \lambda + \epsilon_e(y_e,\hat{y}_e)+\epsilon_h(y_h,\hat{y}_h) \\
    & = \lambda + \rho_e+\rho_h
\end{align}
Given a constant $\lambda$, when the easy-to-adapt subdomain is mostly correct, i.e., $\rho_e \approx 0$, the pseudo risk is bounded by the pseudo rate of the noisy set $\rho_h$.

\subsection*{Hyper-parameter Settings}
We show the hyper-parameters utilized in our experiments in Table~\ref{tab:hyparams}, including $\tau$ for domain division, $\alpha$ for MixUp, $\lambda_{mse}$ that controls the weight of $\mathcal{L}_{mse}$ and the sharpening factor $T$. In semi-supervised learning, to prevent the noisy samples to cause error accumulation, we set $\lambda_{mse}$ to be 0. The Mixup follows a Beta distribution with $\alpha=1.0$. The sharpening factor $T=0.5$. We use $\tau=0.8$ for Office-31 and Office-Home. In VisDA-17, since the model may not perform confidently for the large challenging dataset, we set $\tau=0.5$ to ensure sufficient samples in the easy-to-adapt subdomain. 

\begin{table*}[ht]
	\centering
	\resizebox{0.6\textwidth}{!}{  
	\begin{tabular}{c|ccc} 
		\toprule
			Hyper-parameter\textbackslash Dataset & Office-Home & Office-31 & VisDA-17\\
			\midrule
			$\tau$ & 0.8 & 0.8 & 0.5 \\
			\midrule
			$\alpha$ & 1.0 & 1.0 & 1.0 \\
			\midrule
			$\lambda_{mse}$ & 0. & 0. & 0. \\
			\midrule
			$T$ & 0.5 & 0.5 & 0.5 \\
		\bottomrule
    \end{tabular}
    }
	\caption{Hyper-parameters for different datasets.}
	\label{tab:hyparams}
\end{table*}

\subsection*{Convergence of Losses}
\begin{figure*}[!h]
	\centering
	\includegraphics[width=0.6\textwidth]{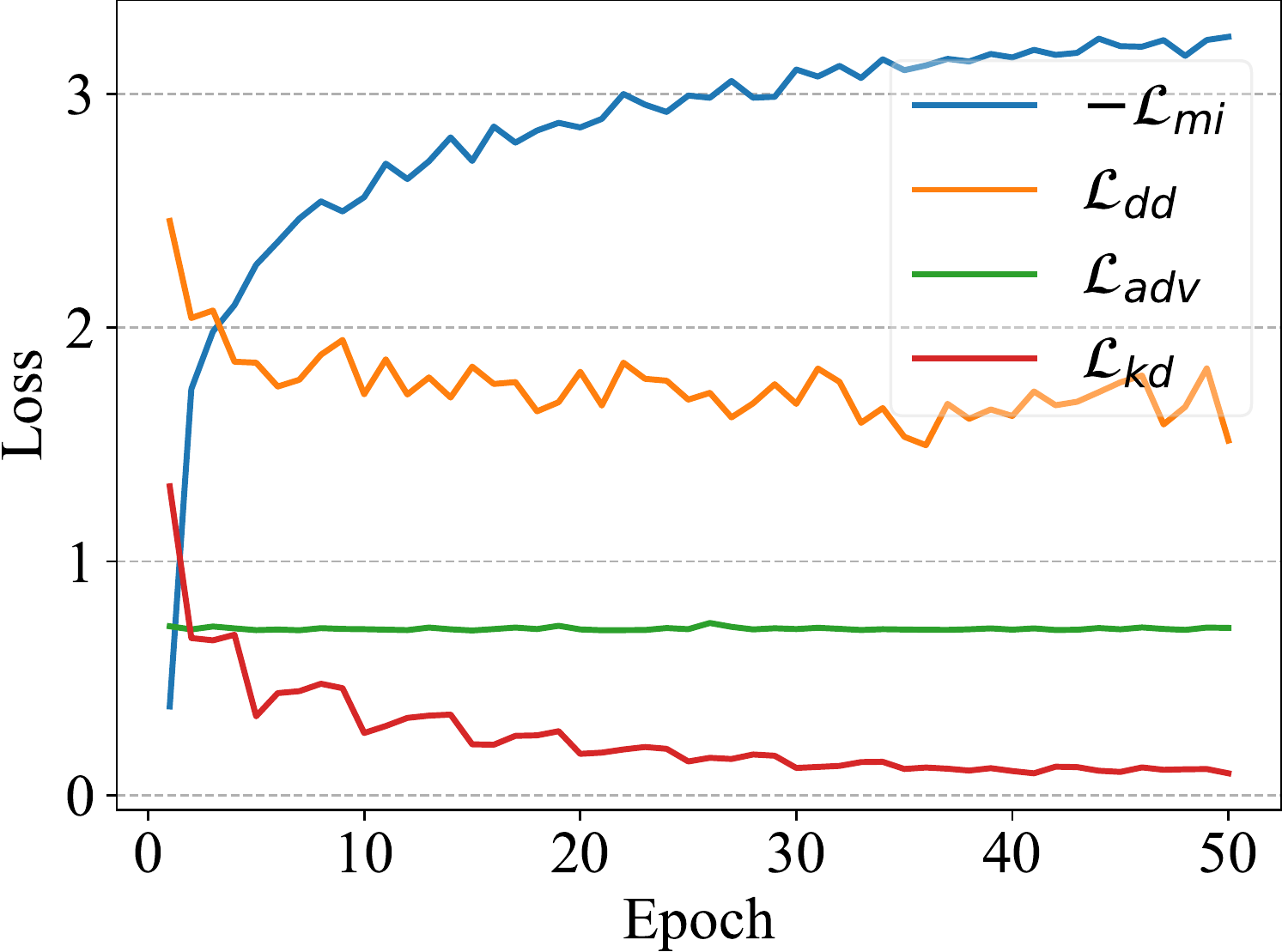}
	\caption{The training procedure of the method.}
	\label{fig:convergence}
\end{figure*}

Figure~\ref{fig:convergence} shows the convergence of the losses of BETA during the training procedure. The adversarial loss keeps small since the two subdomains are all drawn from the same domain and thus the distribution divergence between the two subdomains should be small. The mutual information is maximized as shown in the curve of $\mathcal{L}_{mi}$. The semi-supervised loss $\mathcal{L}_{dd}$ fluctuates while decreasing since two networks utilize the subdomains obtained by each other for semi-supervised learning, which decreases error accumulation.

\subsection*{Domain Division}
\begin{figure}[!h]
	\centering
	\subfigure[Source domain (Art)]{\includegraphics[width=0.95\textwidth, angle=0]{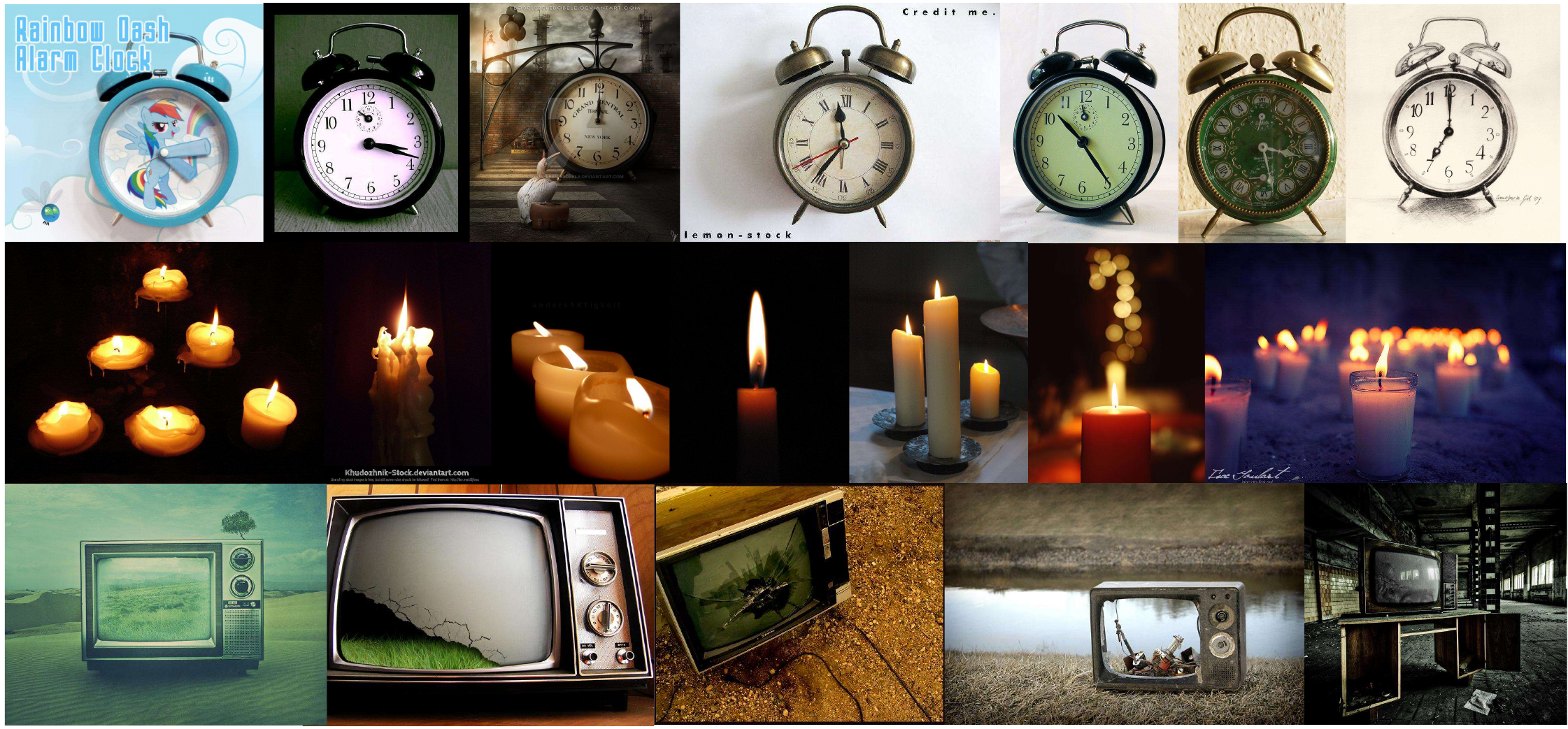}\label{fig:src-domain}}
	\subfigure[Easy-to-adapt subdomain (Clipart)]{\includegraphics[width=0.95\textwidth, angle=0]{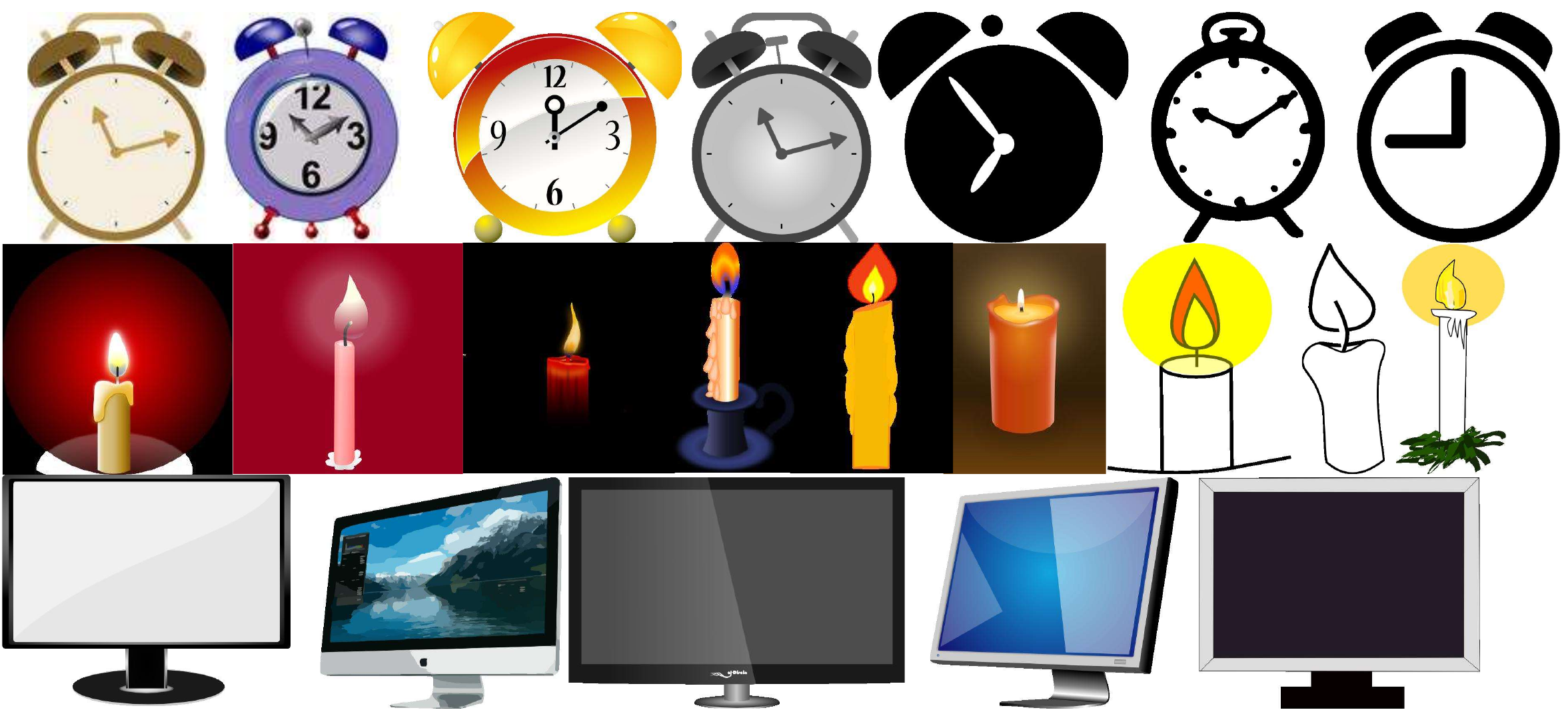}\label{fig:easy-subdomain}}
	\subfigure[Hard-to-adapt subdomain (Clipart)]{\includegraphics[width=0.95\textwidth, angle=0]{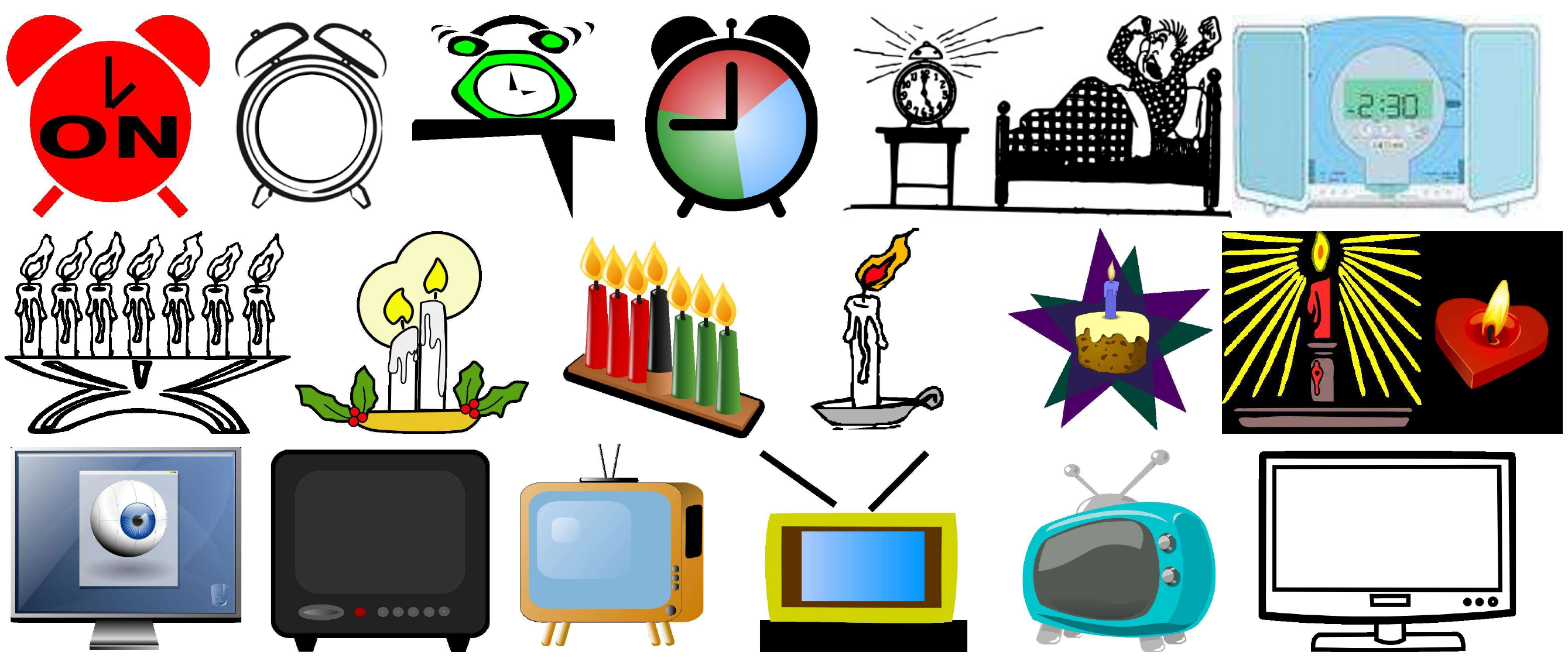}\label{fig:hard-subdomain}}
	\caption{The domain division results on Office-Home (Art$\to$Clipart).}\label{fig:division}
	\vspace{-3mm}
\end{figure}

In Figure~\ref{fig:division}, we show the domain division results at the first epoch (after the warm-up) on Office-Home (Art$\to$Clipart). The three rows contain three categories: alarm clocks, candles, and TV (monitors). The domain shift is very large between \textit{Art} and \textit{Clipart}, and the source-only accuracy is only 44.1\%. Even so, the domain division module still accurately divides the clean easy-to-adapt subdomain and the hard-to-adapt subdomain. In the easy-to-adapt subdomain, the contours of objects are similar to those of the source domain, such as the alarm clock. The domain shift between the easy-to-adapt subdomain and the source domain is smaller, as shown in the candle samples with a black background. For the TV, the easy-to-adapt samples have very clear contours and are easy to recognize. In comparison, the hard-to-adapt subdomain is more challenging in terms of shape, color, and style. Our domain division strategy outputs an AUC of 0.814 for the binary classification of clean samples and noisy samples whose pseudo labels are generated by the source-only model, which enables the semi-supervised learning in BETA to be reasonable. During the training, the AUC keeps increasing to 0.828 and further mitigates the confirmation bias progressively.

\end{document}